\documentclass[journal]{IEEEtran}
\usepackage{cite}
\usepackage{amsmath,amssymb,amsfonts}
\usepackage{algorithmic}
\usepackage{graphicx}
\usepackage{subfigure}
\usepackage{textcomp}
\usepackage[colorlinks,linkcolor=blue]{hyperref}
\usepackage[hyphenbreaks]{breakurl}
\usepackage{multirow}
\usepackage{flushend}

\ifCLASSINFOpdf
\else
\fi
\begin{document}
%
\title{PEA265: Perceptual Assessment of Video Compression Artifacts}
%
%
%

\author{Liqun~Lin,
        Shiqi~Yu,
        Tiesong~Zhao,~\IEEEmembership{Member,~IEEE}
        and~Zhou~Wang,~\IEEEmembership{Fellow,~IEEE}
\thanks{This research is supported by the National Natural Science Foundation of
        China (Grant 61671152).}
\thanks{L. Lin, S. Yu and T. Zhao are with the College of Physics and Information Engineering, Fuzhou University, Fuzhou, Fujian 350116, China. e-mails:\{lin\_liqun, n171120077, t.zhao\}@fzu.edu.cn. Corresponding author: Tiesong Zhao.}
\thanks{Z. Wang is with the Department of Electrical and Computer Engineering, University of Waterloo, Waterloo, ON, Canada. e-mail:
zhou.wang@uwaterloo.ca.}
        }
\maketitle

\begin{abstract}
The most widely used video encoders share a common hybrid coding framework that includes block-based motion estimation/compensation and block-based transform coding. Despite their high coding efficiency, the encoded videos often exhibit visually annoying artifacts, denoted as Perceivable Encoding Artifacts (PEAs), which significantly degrade the visual Quality-of-Experience (QoE) of end users. To monitor and improve visual QoE, it is crucial to develop subjective and objective measures that can identify and quantify various types of PEAs. In this work, we make the first attempt to build a large-scale subject-labelled database composed of H.265/HEVC compressed videos containing various PEAs. The database, namely the PEA265 database, includes 4 types of spatial PEAs ({\it i.e.} blurring, blocking, ringing and color bleeding) and 2 types of temporal PEAs ({\it i.e.} flickering and floating). Each containing at least 60,000 image or video patches with positive and negative labels. To objectively identify these PEAs, we train Convolutional Neural Networks (CNNs) using the PEA265 database. It appears that state-of-the-art ResNeXt is capable of identifying each type of PEAs with high accuracy. Furthermore, we define PEA pattern and PEA intensity measures to quantify PEA levels of compressed video sequence. We believe that the PEA265 database and our findings will benefit the future development of video quality assessment methods and perceptually motivated video encoders.
\end{abstract}


%
\IEEEpeerreviewmaketitle

\section{Introduction}
%
%
%
%
\IEEEPARstart{T}{he} last decade has witnessed a booming of High Definition (HD)/Ultra HD (UHD) and 3D/360-degree videos due to the rapid developments of video capturing, transmission and display technologies. According to Cisco Visual Networking Index (VNI)\cite{1}, video content has taken over 2/3 bandwidth of current broadband and mobile networks, and will grow to 80\%-90\% in the visible future. To meet such a demand, it is necessary to improve network bandwidth and maximize video quality under a limited bitrate or bandwidth constraint, where the latter is generally achieved by lossy video coding technologies.

The widely used video coding schemes are lossy for two reasons. Firstly, Shannon's theorem sets the limit of lossless coding, which cannot fulfill the practical needs on video compression; secondly, the Human Vision System (HVS)\cite{2} is not uniformly sensitive to visual signals at all frequencies, which allows us to suppress certain frequencies with negligible loss of perceptual quality. State-of-the-art video coding schemes, such as H.264 Advanced Video Coding (H.264/AVC) \cite{3}, H.265 High Efficiency Video Coding (H.265/HEVC)\cite{4}, Google VP8/VP9\cite{5,6}, China's Audio-Video coding Standards (AVS/AVS2)\cite{7,8}, adopt the conventional hybrid video coding structure. This infrastructure, originated from 1980s\cite{9}, consists of a group of standard procedures including intra-frame prediction, inter-frame motion estimation and compensation, followed by spatial transmission, quantization and entropy coding. To facilitate these functions in videos of large sizes, the encoder further divides the frames into slices and coding units. Thereby, when the bitrate is not sufficially high, the compressed video encompasses various types of information loss within and across blocks, slices and units, resulting in visually unnatural structure impairments or perceptual artifacts \cite{10}. These Perceivable Encoding Artifacts (PEAs) greatly degrade the visual Quality-of-Experience (QoE) of users \cite{11,12}.

 The detection and classification of PEAs are challenging tasks. In video encoders, conventional quality metrics such as Sum of Absoluted Differences (SAD) \cite{13}, Sum of Squared Errors (SSE) \cite{14}, Peak-Signal-to-Noise Ratio (PSNR) \cite{15},  and Structural SIMilarity (SSIM) index \cite{15} are weak indicators of PEAs. At the user-end, the PEAs are highly visible but not properly measured. Recent developments have greatly put forward the 4K/8K era and  user-centric video coding and delivery has become ever important \cite{16}. Meanwhile, the advancement of computing and networking technologies have enabled deep investigations on PEA recognition and quantification.
\begin{figure*}[!ht]
  \centering
  \subfigure[Reference frame]{
    \label{fig_S_Blurring1}
    \includegraphics[width=3.4in,height=3.5cm]{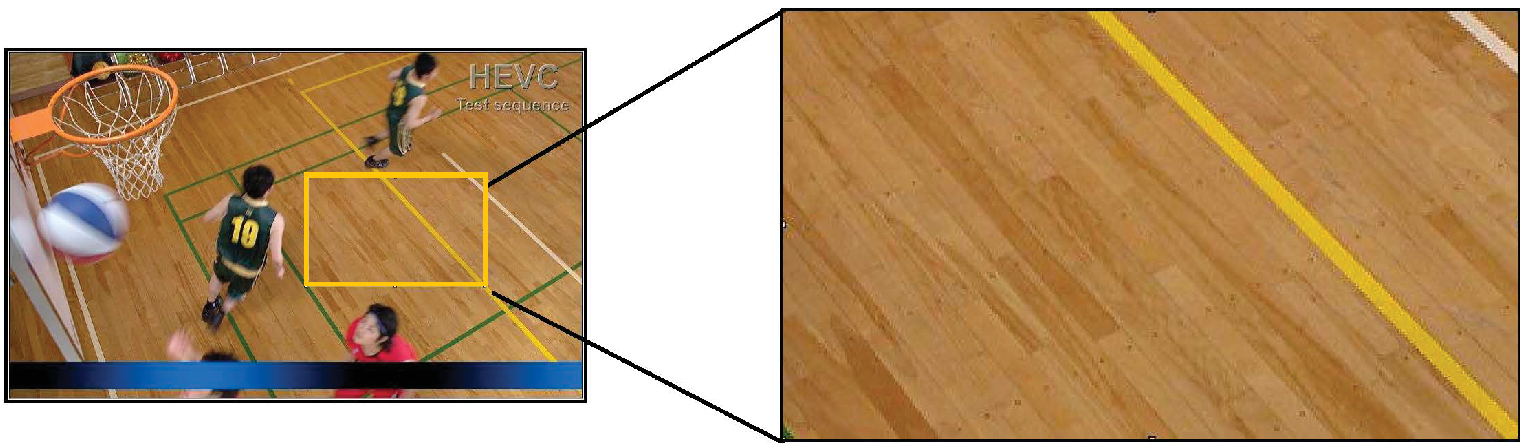}
  }
  \subfigure[Compressed frame with blurring artifact]{
    \label{fig_S_Blurring2}
    \includegraphics[width=3.4in,height=3.5cm]{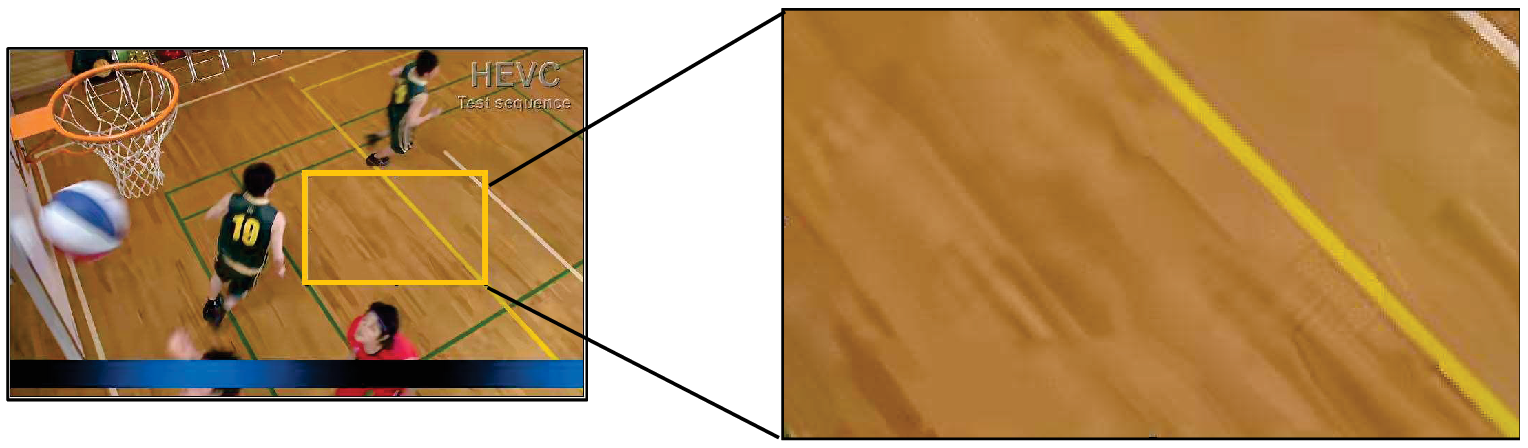}
  }
  \caption{An example of blurring artifact.}
  \label{fig_blurring}
\end{figure*}

In\cite{17}, the classification of diversified PEAs have been elaborated. In \cite{18}, it is observed that these PEAs have significant impacts on visual quality of H.264/AVC. Specifically, 96\% of quality variance could be predicted by the intensities of three common PEAs: blurring, blocking and color bleeding. Until now, blocking and blurring artifacts have been extensively investigated, which are caused by spatial-inconsistent and high-frequency signal losses  respectively. In many hybrid encoders, de-blocking filters are introduced to prevent severe blocking artifacts, which may, however, introduce high blurriness \cite{19}. Other typical artifacts, such as ringing \cite{20} and color bleeding \cite{21}, may be generated due to errors in high frequencies of luma and chroma signals, respectively. To address these issues, intricate schemes have been developed to PEA removal \cite{22,23,24}. However, due to their high complexities, these algorithms are usually deployed at the post-processing stage instead of video compression. Meanwhile, temporal PEAs have also attracted significant attention. In \cite{25}, a simplified robust statistical model and the Huber statistical model for temporal artifact reduction are proposed. Gong {\it et al.} \cite{26} presented the hierarchical prediction structure to find plausible reasons of temporal artifacts. Meanwhile, a metric for just noticeable temporal artifact and an efficient temporal PEA eliminating algorithm in video coding were proposed. In addition, Zeng {\it et al.} \cite{17} presented an algorithm detecting and locating the floating artifacts. Despite these efforts, there is still a lack of subjective and objective approaches to systematic PEA recognition and analysis. Recently, deep learning techniques \cite{27}, especially Convolutional Neural Network (CNN) \cite{28}, have demonstrated their promise in improving video coding performance \cite{29,30,31,32,33}. This inspired us to introduce CNN to the recognition of PEAs in hybrid encoding.
\begin{figure*}[!ht]
  \centering
  \subfigure[Reference frame]{
    \label{fig_S_Blocking1}
    \includegraphics[width=3.4in,height=3.5cm]{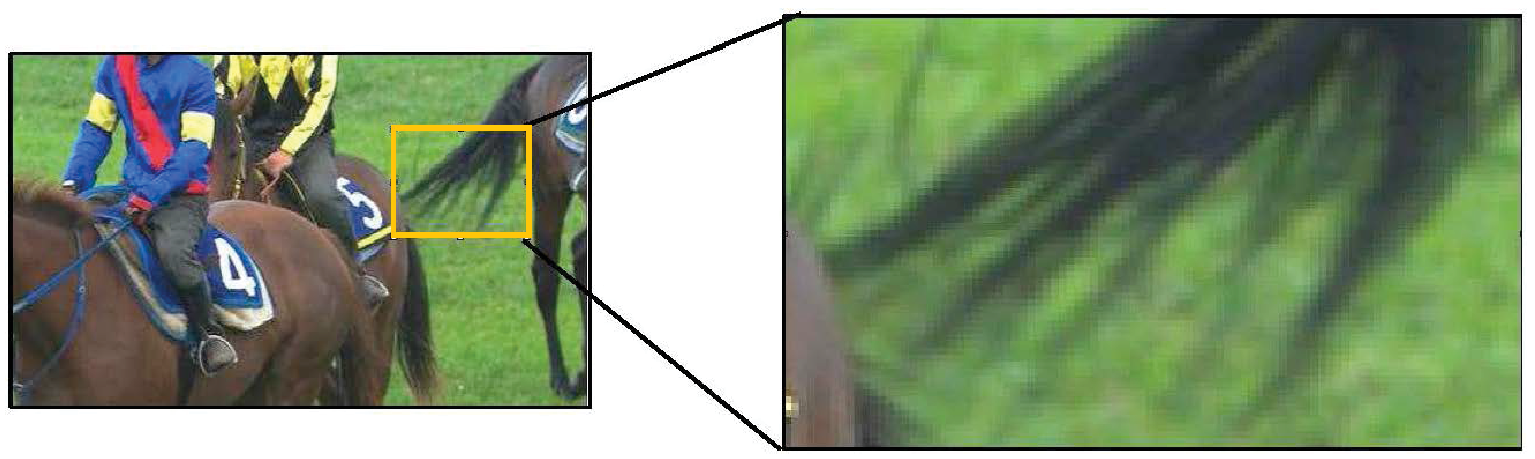}
  }
  \subfigure[Compressed frame with blocking artifact]{
    \label{fig_S_Blocking2}
    \includegraphics[width=3.4in,height=3.5cm]{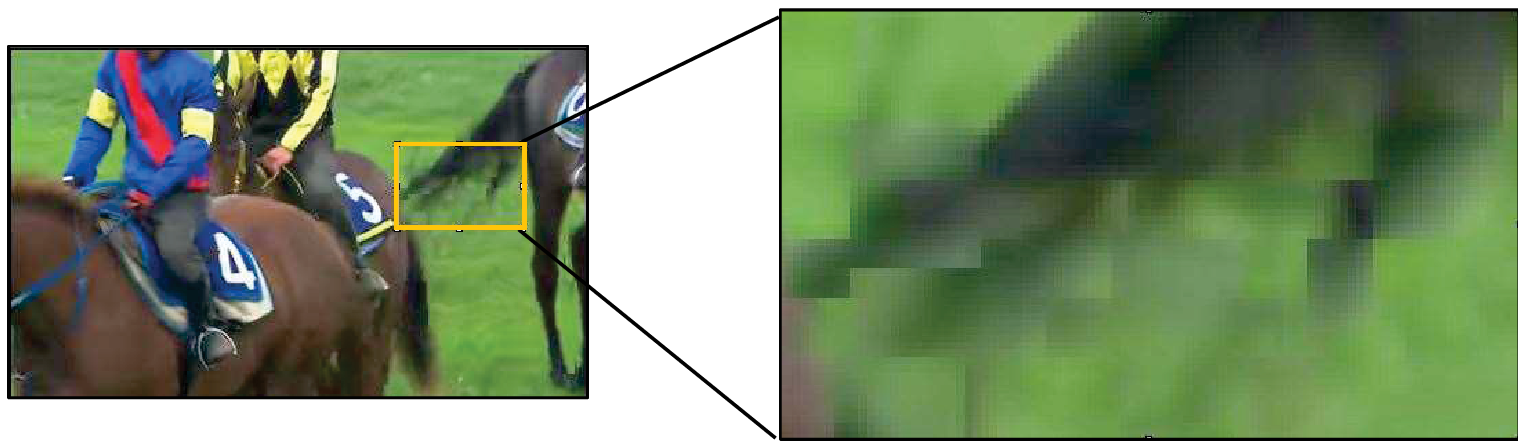}
  }
  \caption{An example of blocking artifact.}
  \label{fig_blocking}
\end{figure*}
 In this work, we employ state-of-the-art video encoder H.265/HEVC to develop a PEA database, namely the PEA265 database, for PEA recognition. The contributions of this work are as follows:

(1) A subjective-labelled database of compressed videos with PEAs. We select 6 typical PEAs based on \cite{17}. We utilize the H.265/HEVC to encode a group of standard sequences and recruit users to mark all the 6 types of PEAs. Finally, we cut the marked sequences into image/video patches with positive and negative PEA labels. In total, there are 6 typical PEAs and at least 60,000 positive or negative labels are given for each type of PEA.

(2) An objective PEA recognition approach based on CNN. For each type of PEA, we construct and compare LeNet \cite{34} and ResNeXt \cite{35} to recognize PEA types. It appears that state-of-the-art ResNeXt outperforms LeNet in terms of PEA recognition. We are able to achieve an accuracy of at least 80\% for all PEAs types.

(3) A PEA intensity measure for a compressed video sequence. By summarizing all PEA recognitions, we obtain an overall PEA intensity measure of a compressed video sequence, which helps characterize the subjective annoyance of PEAs in compressed video.
\begin{figure*}[!ht]
  \centering
  \subfigure[Reference frame]{
    \label{fig_S_Ring1}
    \includegraphics[width=3.4in,height=3.5cm]{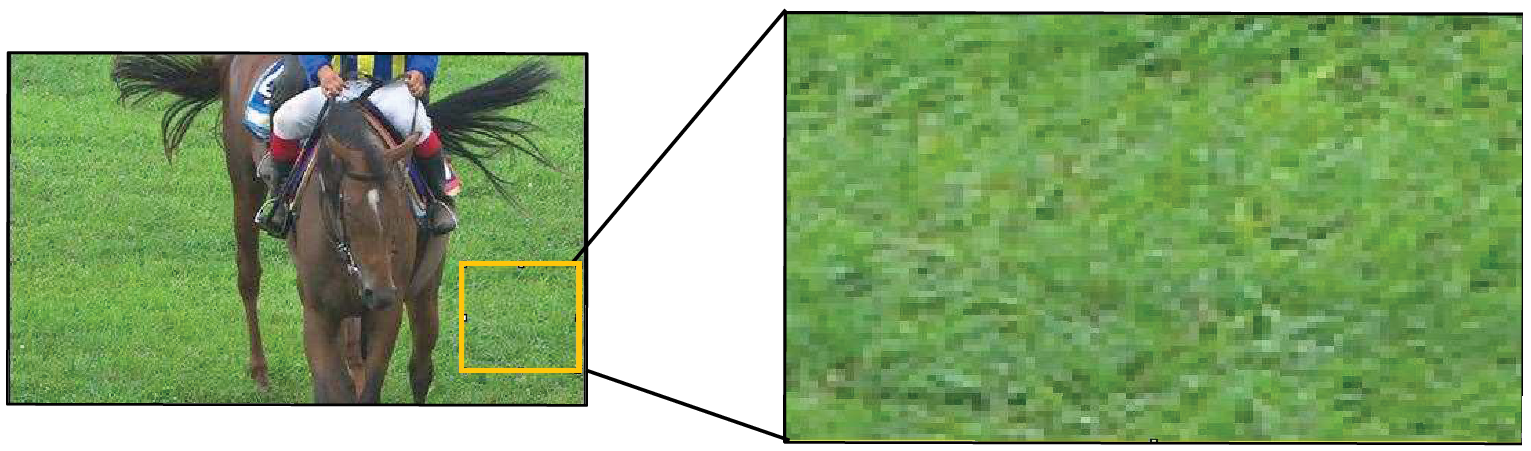}
  }
  \subfigure[Compressed frame with ringing artifact]{
    \label{fig_S_Ring2}
    \includegraphics[width=3.4in,height=3.5cm]{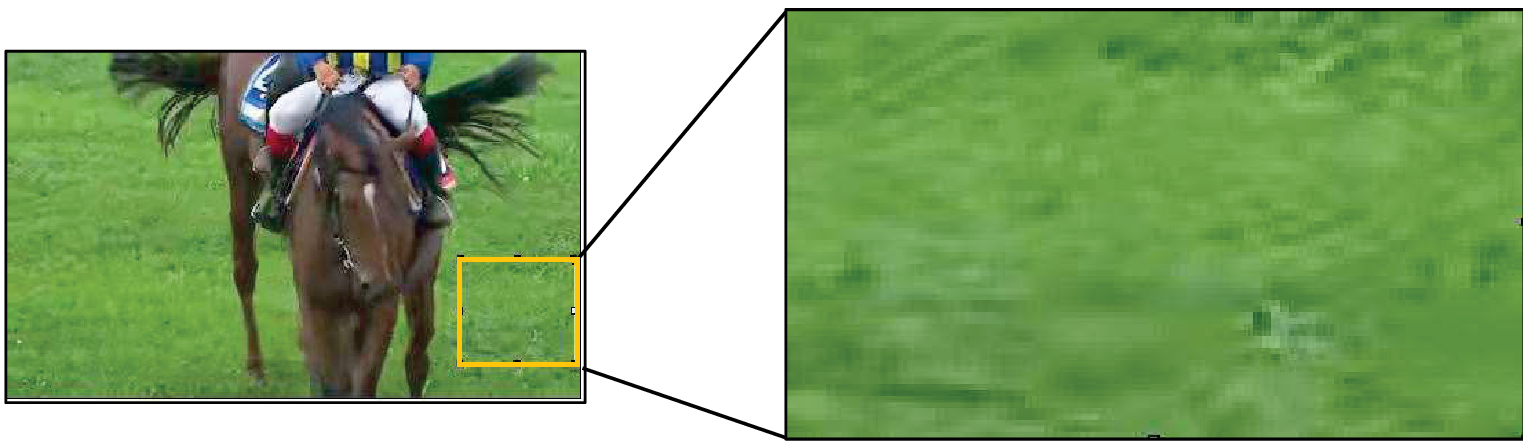}
  }
  \caption{An example of ringing artifact.}
  \label{fig_ringing}
\end{figure*}
\begin{figure*}[!ht]
  \centering
  \subfigure[Reference frame]{
    \label{fig_S_Bleeding1}
    \includegraphics[width=3.4in,height=3.5cm]{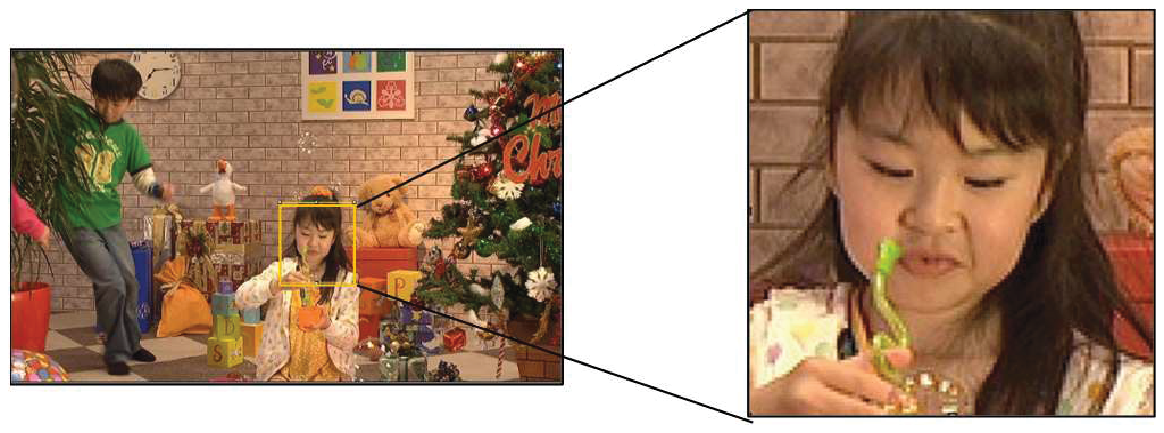}
  }
  \subfigure[Compressed frame with color bleeding artifact]{
    \label{fig_S_Bleeding2}
    \includegraphics[width=3.4in,height=3.5cm]{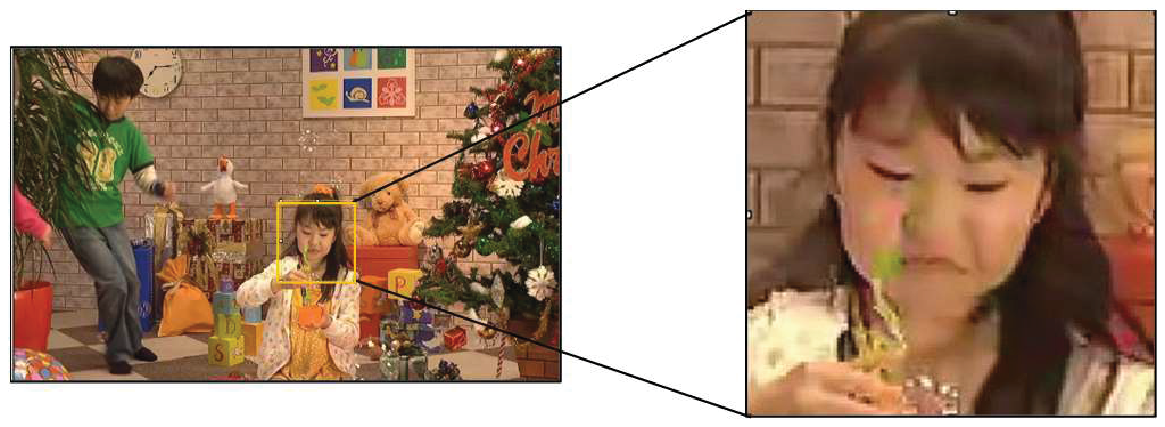}
  }
  \caption{An example of color bleeding artifact.}
  \label{fig_bleeding}
\end{figure*}

The rest of the paper is organized as follows. In Section II, we discuss diversified PEAs in H.265/HEVC and select 6 types of PEAs to develop our database. In Section III, we elaborate the details of our subjective database including video sequence preparation, subjective testing and data processing. Section IV presents our deep learning-based  PEA recognition and the overall PEA intensity measurement. Finally, Section V concludes the paper.



\section{PEA Classification}
In this section, we review the PEA classification in \cite{17} and select typical PEAs to develop our subjective database. According to \cite{17}, the PEAs are classified into spatial and temporal artifacts, where spatial artifacts include blurring, blocking, color bleeding, ringing and basis pattern effect; temporal artifacts include floating, jerkiness and flickering. In this work, we select blurring, blocking, color bleeding, ringing of spatial artifacts and floating, flickering of temporal artifacts in the development of our database. Basis pattern effect and jerkiness artifacts are excluded because: 1) the basis pattern effect has similar visual appearance and has similar origin to the ringing effect; 2) the jerkiness artifacts are caused by image capturing factors such as frame rate instead of compression. We summarize the characteristics and plausible reasons of the 6 typical types of PEAs as follows.

\subsection{Spatial Artifacts}
Block-based video coding schemes create various spatial artifacts due to block partitioning and quantization. The spatial artifacts, with different visual appearances, can be identified without temporal reference.

\subsubsection{Blurring}
Aiming at a higher compression ratio, the HEVC encoder quantizes transformed residuals discrepantly. When the video signals are reconstructed, high frequency energy may be severely lost, which may lead to visual blur. Perceptually, blurring usually appears as the loss of spatial details or sharpness of edges or texture regions in an image. An example is shown in the marked rectangular region in Fig. \ref{fig_blurring} (b). It displays the spatial loss of the basketball field.

\begin{figure*}[!ht]
  \centering
  \subfigure[Reference frame]{
    \label{fig_S_Flickering1}
    \includegraphics[width=3.4in,height=3.5cm]{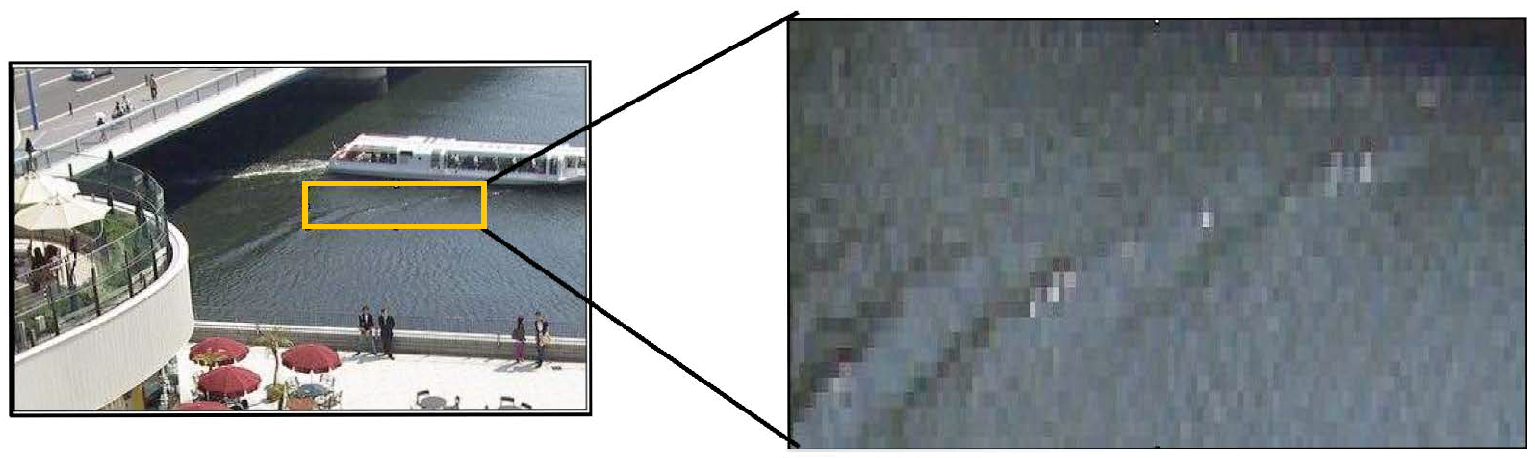}
  }
  \subfigure[Compressed frame with flickering artifact]{
    \label{fig_S_Flickering2}
    \includegraphics[width=3.4in,height=3.5cm]{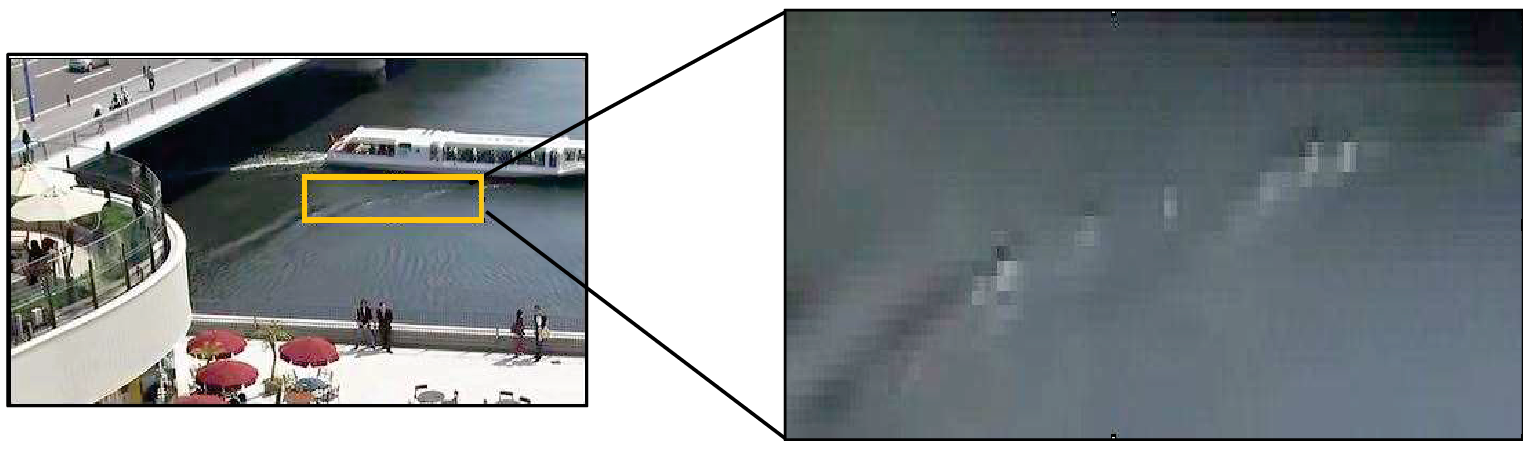}
  }
  \caption{An example of flickering artifact.}
   \label{fig_flickering}
\end{figure*}

\begin{figure*}[!ht]
  \centering
  \subfigure[Reference frame]{
    \label{fig_S_Floating1}
    \includegraphics[width=3.4in,height=3.5cm]{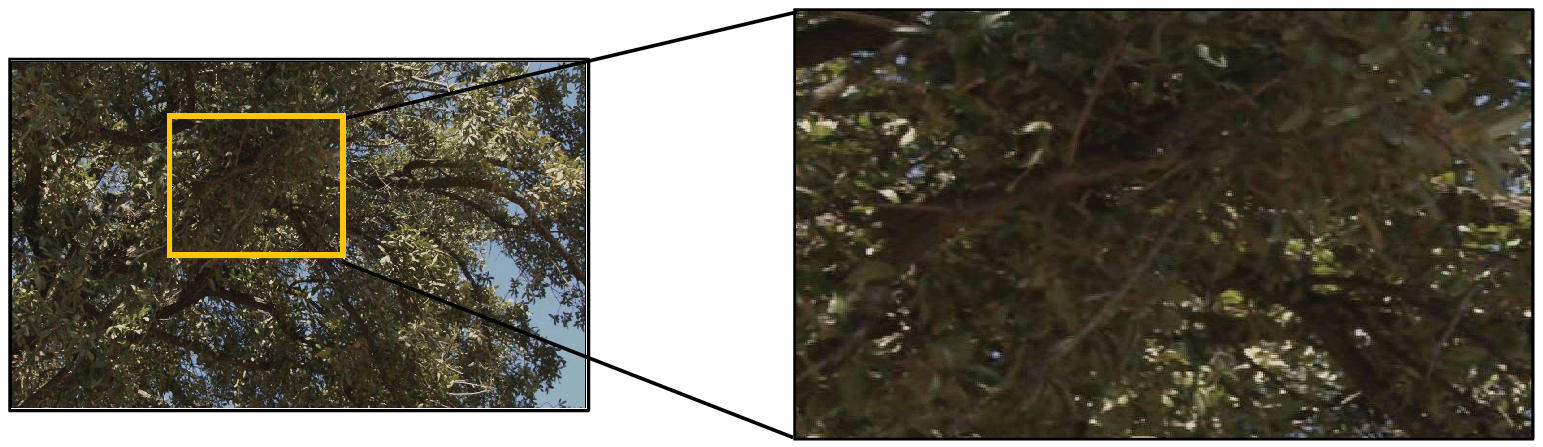}
  }
  \subfigure[Compressed frame with floating artifact]{
    \label{fig_S_Floating2}
    \includegraphics[width=3.4in,height=3.5cm]{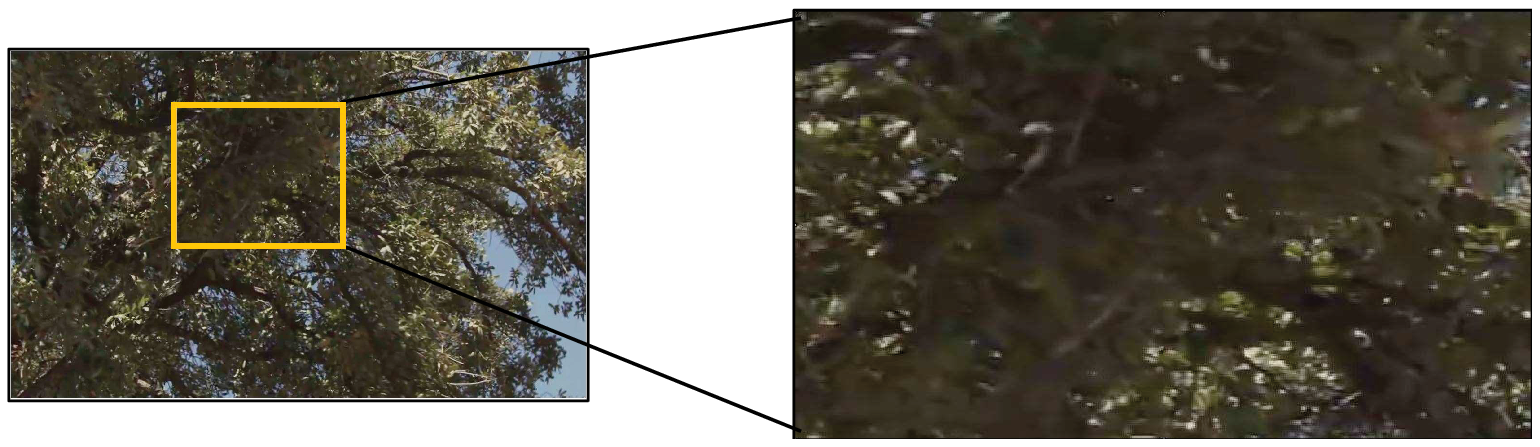}
  }
  \caption{An example of floating artifact.}
   \label{fig_floating}
\end{figure*}

\subsubsection{Blocking}
The HEVC encoder is block-based, and all compression processes are performed within non-overlapped blocks. This often results in false discontinuities across block boundaries. The visual appearance of blocking may be different subject to the region of visual discontinuities. In Fig. \ref{fig_blocking} (b), a blocking example of the horse tail is highlighted in the marked rectangular region.

\begin{table*}[!ht]
\renewcommand{\arraystretch}{1.3}
\caption{Testing sequences}
\label{table_sequence}
\centering
\begin{tabular}{|c|c|c|c|c|c|c|c|c|c|}
\hline
No & Class & Sequence (Resolution)   & Frames & Frame rate & No & Class & Sequence (Resolution)    & Frames  & Frame rate \\ \hline
1     & A & \emph{Traffic} (2560x1600)         & 150         & 30fps      & 13     & C & \emph{BasketballDrill} (832x480)    & 500         & 50fps      \\ \hline
2     & A & \emph{PeopleOnStreet} (2560x1600)  & 150         & 30fps      & 14     & D & \emph{RaceHorses} (416x240)         & 300         & 30fps      \\ \hline
3     & A & \emph{NebutaFestival} (2560x1600)          & 300         & 60fps      & 15   & D & \emph{BQSquare} (416x240)           & 600         & 60fps      \\ \hline
4     & A & \emph{SteamLocomotive} (2560x1600) & 300         & 60fps      & 16  & D & \emph{BlowingBubbles} (416x240)     & 500         & 50fps      \\ \hline
5     & B & \emph{Kimono} (1920x1080)          & 240         & 24fps      & 17   & D & \emph{BasketballPass} (416x240)     & 500         & 50fps      \\ \hline
6     & B & \emph{ParkScene} (1920x1080)       & 240         & 24fps      & 18   & E & \emph{FourPeople} (1280x720)        & 600         & 60fps      \\ \hline
7     & B & \emph{Cactus} (1920x1080)          & 500         & 50fps      & 19    & E & \emph{Johnny} (1280x720)            & 600         & 60fps      \\ \hline
8     & B & \emph{BQTerrace} (1920x1080)       & 600         & 60fps      & 20    & E & \emph{KristenAndSara} (1280x720)    & 600         & 60fps      \\ \hline
9     & B & \emph{BasketballDrive} (1920x1080) & 500         & 50fps      & 21   & F & \emph{BaskeballDrillText} (832x480) & 500         & 50fps      \\ \hline
10     & C & \emph{RaceHorses} (832x480)        & 300         & 30fps     & 22   & F & \emph{SlideEditing} (1280x720)      & 300         & 30fps      \\ \hline
11     & C & \emph{BQMall} (832x480)            & 600         & 60fps     & 23   & F & \emph{SlideShow} (1280x720)         & 500         & 20fps      \\ \hline
12     & C & \emph{PartyScene} (832x480)        & 500         & 50fps      &      &   &                             &             &            \\ \hline
\end{tabular}
\end{table*}
\subsubsection{Ringing}
Ringing is caused by the coarse quantization of high frequency components. When the high frequency component of oscillating structure has a quantization error, the pseudo structure may appear near strong edges (high contrast), which manifests artificial wave-like or ripple structures, denoted as ringing. A ringing example is given in the marked rectangular region in Fig. \ref{fig_ringing} (b).

\subsubsection{Color bleeding}
The chromaticity information is coarsely quantized to cause color bleeding. It is related to the presence of strong chroma variations in the compressed images leading to false color edges. It may be a result of inconsistent image rendering across the luminance and chromatic channels. A color bleeding example is provided in the marked rectangular region in Fig. \ref{fig_bleeding} (b), which exhibits chromatic distortion and additional inconsistent color spreading in the rendering result.

\subsection{Temporal Artifacts}
Temporal artifacts are manifested as temporal information loss, and can be identified during video playback.

\subsubsection{Flickering}
Flickering is usually frequent brightness or color changes along the time dimension. There are different kinds of flickering including mosquito noise, fine-granularity flickering and coarse-granularity flickering. Mosquito noise is high frequency distortion and the embodiment of the coding effect in the time domain. It moves together with the objects like mosquitoes flying around. It may be caused by the mismatch prediction error of the ringing effect and the motion compensation. The most likely cause of coarse-granulating blinking may be luminance variations across Group-Of-Pictures (GOPs). Fine-granularity flickering may be produced by slow motion and blocking effect. An example is given in the marked rectangular region in Fig. \ref{fig_flickering} (b). Frequent luminance changes on the surface of the water produce flickering artifacts.

\subsubsection{Floating}
Floating refers to the appearance of illusory movements in certain areas rather than their surrounding environment. Visually these regions create a strong illusion as if they are floating on top of the surrounding background. Most often, a scene with a large textured area such as water or trees is captured with cameras moving slowly. The floating artifacts may be due to the skip mode in video coding, which simply copies a block from one frame to another without updating the image details further. Fig. \ref{fig_floating} (b) gives a floating example. Visually these regions create a strong illusion as if they are floating on top of the leaves.


\section{PEA265 Database}
The development of the PEA265 database is composed of four steps: preparation of test video sequences, subjective PEA region identification, patch labeling, and formation of PEA265 database.

\subsection{Testing Video Sequences}
The selection of testing sequences follows the Common Test Conditions (CTC) \cite{36}. These standard test sequences in YUV4:2:0 format are summarized in Table \ref{table_sequence}. We employs HEVC encoder \cite{37} to compress the video sequences with four Quantization parameter (Qp) values of 22, 27, 32 and 37, respectively. Four types of coding structures are covered: all intra, random access, low delay and low delay P. Thus, there are totally 320 encoded sequences. For consistency, the output bit depth is set to 8.

\subsection{Subjective PEA Region Identification}
In order to identify all PEAs, we ask subjects ({\it i.e.} testees) to label all video sequences. Our testing procedure follows the ITU-R BT.500 \cite{38} document with two phases. In the pre-training phase, all subjects are told about our testing procedures and trained to identify PEAs. In the formal-testing phase, all subjects are asked to watch these sequences and circle PEA regions. The test sequences are presented in random order. Mid-term breaks are set during the formal-testing to avoid visual fatigue. 30 subjects, 14 males and 16 females, aged between 20 and 22, participated in the subjective experiment.

\begin{figure*}[!ht]
  \centering
  \subfigure[Patch labeling in a compressed video frame]{
    \label{fig_S_Spa1}
    \includegraphics[width=3in,height=3.5cm]{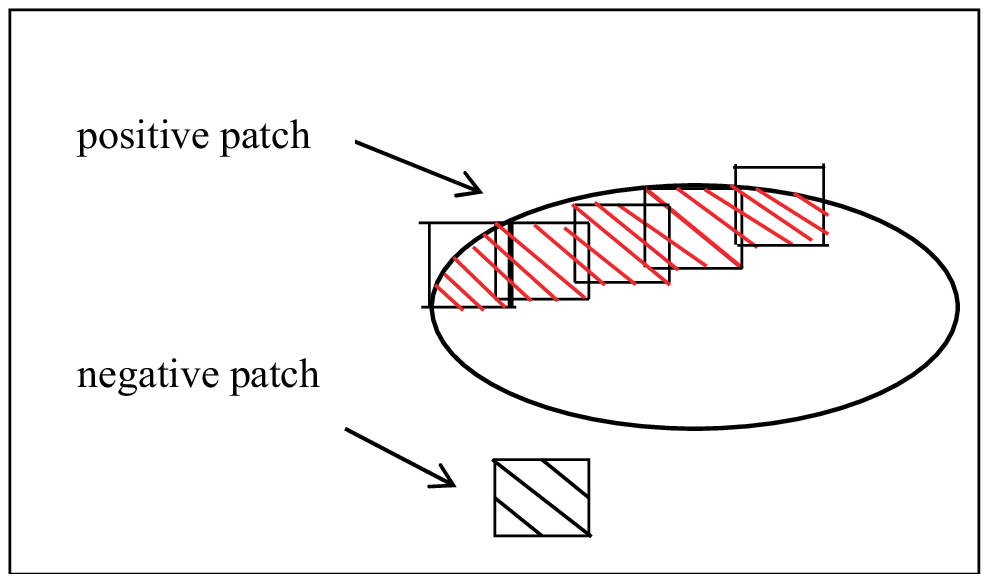}
  }
  \subfigure[Patch labeling in corresponding reference video frame]{
    \label{fig_S_Spa2}
    \includegraphics[width=3in,height=3.5cm]{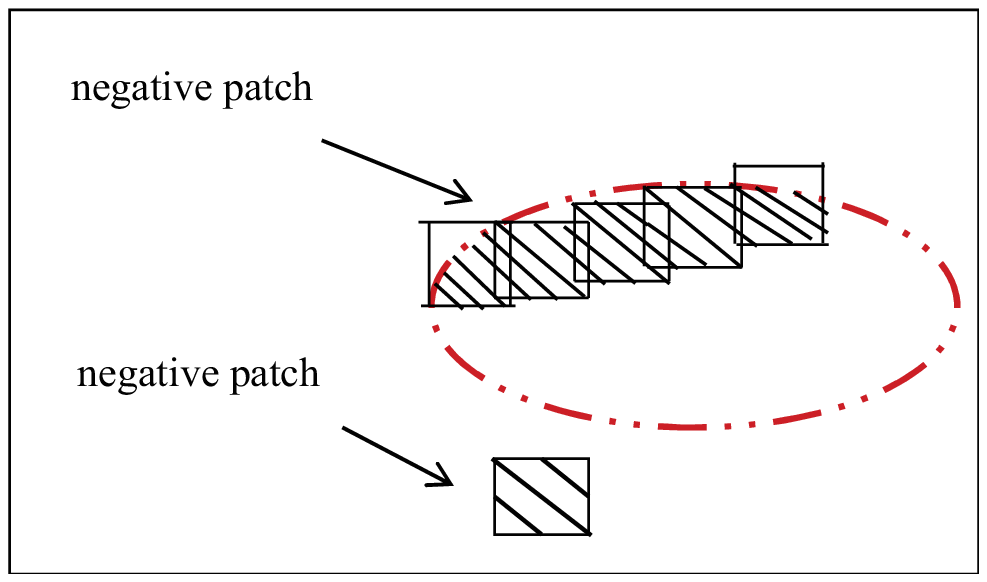}
  }
  \caption{ Positive/negative patch labeling for spatial PEAs.}
  \label{fig_spatial}
\end{figure*}

\begin{figure*}[!ht]
  \centering
  \subfigure[Patch labeling in compressed video frames]{
    \includegraphics[width=3in,height=3.5cm]{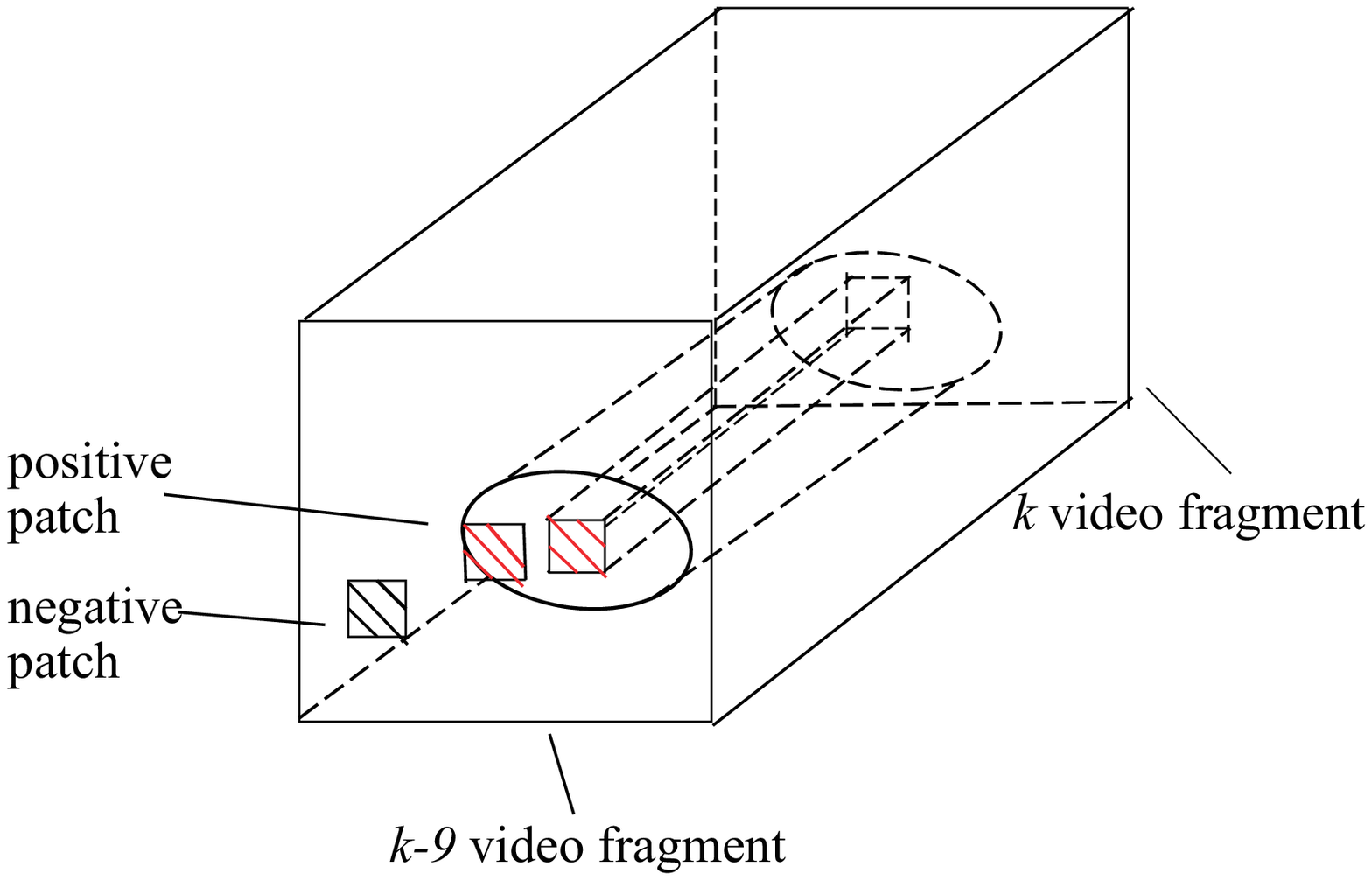}
  }
  \subfigure[Patch labeling in corresponding reference video frames]{
    \includegraphics[width=3in,height=3.5cm]{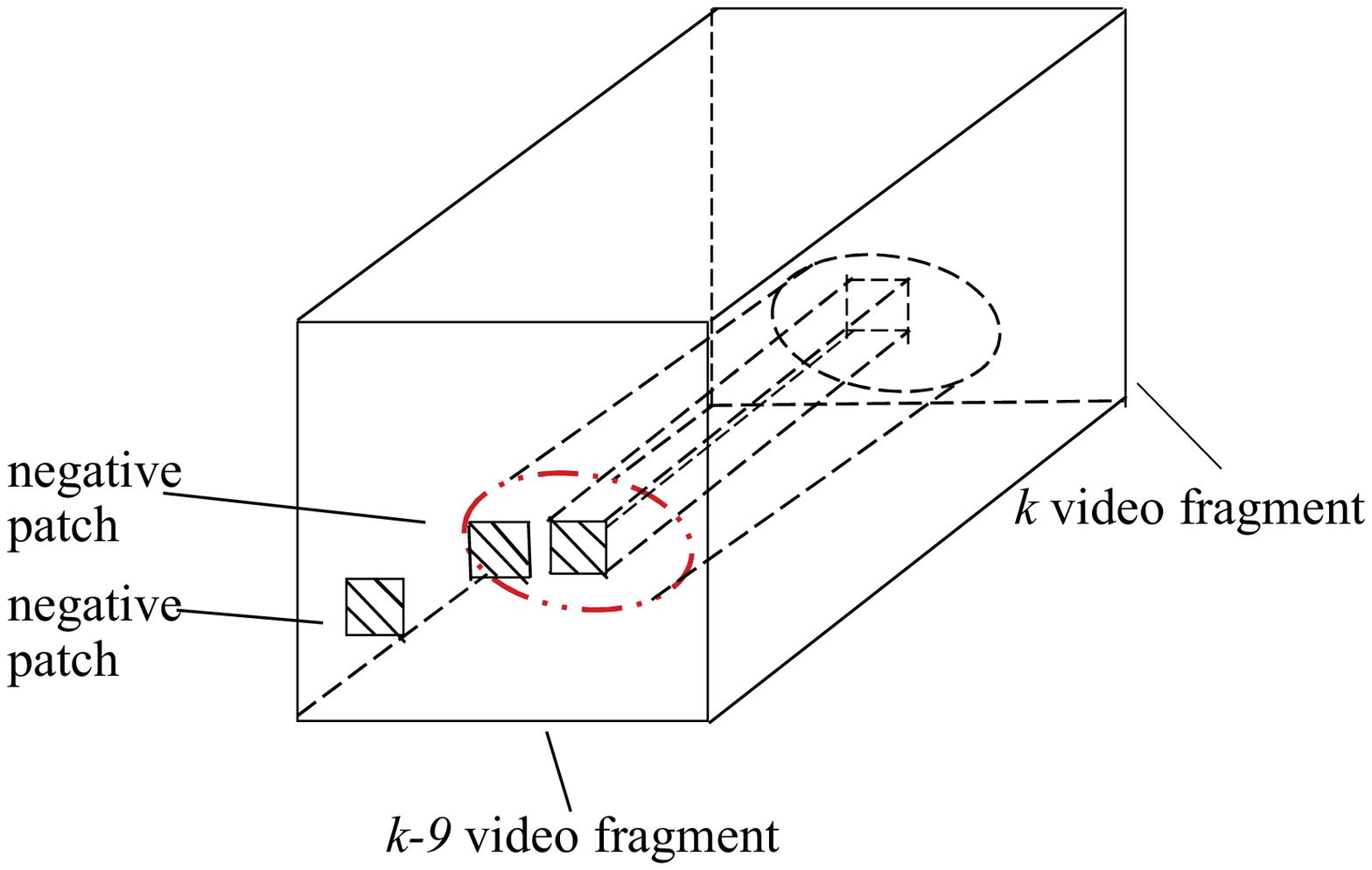}
  }
  \caption{Positive/negative patch labeling for temporal PEAs.}
  \label{fig_temporal}
\end{figure*}

\subsection{Patch Labeling}
During subjective test, the PEA regions are circled by subjects (may be an ellipse shape) and saved in binary files, from which, we derive positive and negative patches in rectangular or cuboid shapes.

\subsubsection{Spatial artifacts}
For spatial artifacts, we label the patches by a sliding window of 32$\times$32 or 72$\times$72. In a compressed video, if at least half of the pixels within the sliding window belong to this circled region, it is labeled as positive; otherwise negative. Patches belonging to the corresponding frame of uncompressed video are randomly selected and categorized as negative, whether or not they are co-located within the circled region. The ratio between the numbers of the two types of negative patches is 1:2. The labeling process is illustrated in Fig. \ref{fig_spatial}.

\subsubsection{Temporal artifacts}
Temporal PEAs appear in a group of successive video frames. When a testee pauses video playback and marks a temporal artifact region, 10 frames starting from the current frames are extracted. The video fragment is then further checked by a spatial sliding window of 32$\times$32 or 72$\times$72: if at least half of the pixels in this window are within the circled region, then the corresponding cuboid is labelled as positive, otherwise negative. Similar to spatial artifacts, negative temporal patches are also obtained from co-located region in the uncompressed sequences. This process is illustrated in Fig. \ref{fig_temporal}.
\begin{figure*}[!t]
\centering
\includegraphics[width=6.4in,height=5cm]{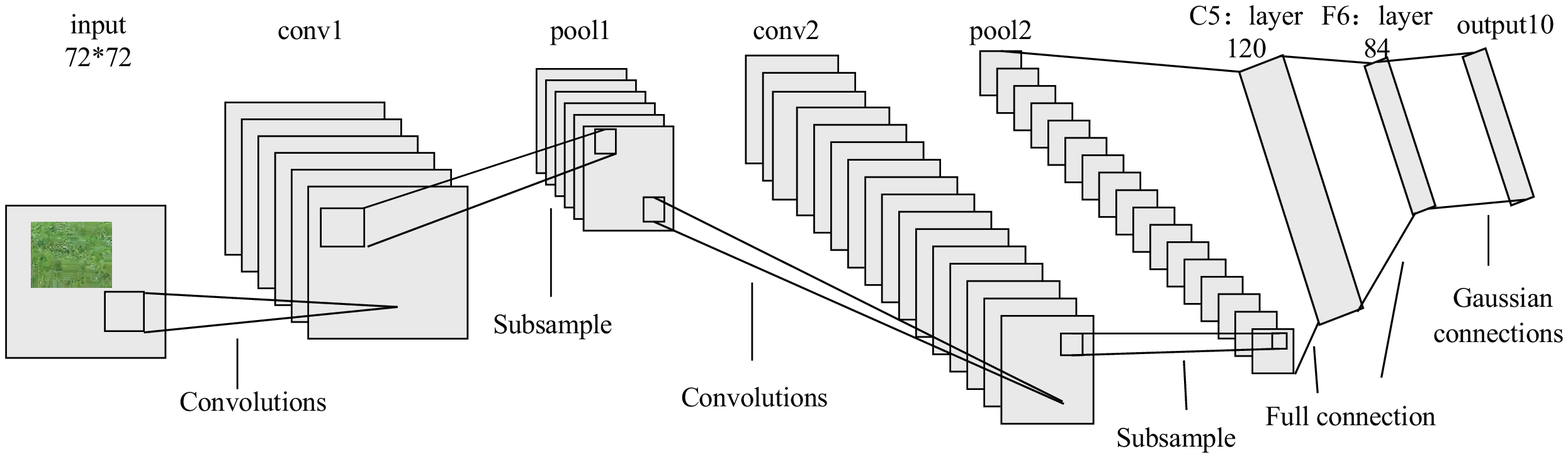}
\caption{The LeNet-5 structure.}
\label{fig_LeNet_architecture}
\end{figure*}
\subsection{Summary of the database}
The PEA265 database covers 6 types of PEAs including 4 types of spatial PEAs (blurring, blocking, ringing and color bleeding) and 2 types of temporal PEAs (flickering and floating). Each type of PEAs contains at least 60,000 image or video patches with positive and negative labels, respectively. Three typical PEA (ringing, color bleeding and flickering) patches are of
 size 32$\times$32, and the other two (blurring, blocking and floating) are of size 72$\times$72. These patches are stored in binary format. The total data size is about 28Gb. Each PEA patch, is indexed by its video name, frame number, and coordinate position.

\section{CNN-based PEA Recognition}
In this section, we utilize the PEA265 database to train a deep-learning-based PEA recognition model. We also propose two metrics, PEA pattern and PEA intensity, which can be further employed in vision-based video processing and coding.

\subsection{Subjective recognition with CNN}

We choose two popular CNN architectures, LeNet \cite{34} and ResNeXt \cite{35} in this study. For each type of PEA, we randomly select 50,000 ground-truth samples from PEA265 database. These samples are further split to 75:25 training/testing sets.
\begin{table}[!t]
\renewcommand{\arraystretch}{1.3}
\caption{Training/Testing Recognition Accuracy sets.}
\label{table_accuary}
\centering

\begin{tabular}{|c|c|c|c|c|}
\hline
\multirow{2}{*}{PEAs}
               & \multicolumn{2}{c|}{LeNet-5} & \multicolumn{2}{c|}{ResNeXt} \\ \cline{2-5}
     & Training      & Testing      & Training      & Testing      \\ \hline
Blurring       & 0.6833        & 0.6768       & 0.9352        & 0.8176       \\ \hline
Blocking       & 0.7154        & 0.7162       & 0.9514        & 0.9281       \\ \hline
Ringing        & 0.6946        & 0.6917       & 0.8524        & 0.8356       \\ \hline
Color bleeding & 0.7172        & 0.7200       & 0.8706        & 0.8494       \\ \hline
Flickering     & 0.6572        & 0.6496       & 0.8108        & 0.8019       \\ \hline
Floating       & 0.7096        & 0.7087       & 0.8228        & 0.8051       \\ \hline
\end{tabular}
\end{table}
\subsubsection{LeNet-5 network}
The LetNet architecture is a classic classifier CNN. In our work, We use eight layers (including input) with its structure given in Fig. \ref{fig_LeNet_architecture}. The conv1 layer learns 20 convolution filters of size 5$\times$5. We apply a ReLU activation function followed by 2$\times$2 max-pooling in both x$\times$y direction with a stride of 2. The conv2 layer learns 50 convolution filters. Finally, the softmax classifier is applied to return a list of probabilities. The class label with the largest probability is chosen as the final classification from the network. Here, the input samples are of sizes 32$\times$32 or 72$\times$72, and are in binary format. In order to obtain a higher accuracy, we augment the training data by rotation, width scaling, height scaling, shear, zoom, horizontal flip and fill mode. After data augmentation, the accuracy improves by about 10\% to 70\% as shown in Table \ref{table_accuary}.

\begin{figure}[!t]
\centering
\includegraphics[width=3.2in,height=4cm]{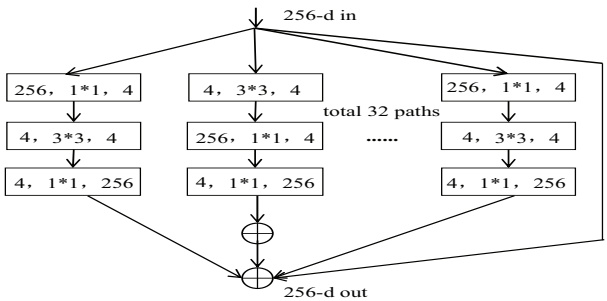}
\caption{A block of ResNeXt with cardinality = 32.}
\label{fig_resNet_architecture}
\end{figure}

\subsubsection{ResNeXt network}
 The ResNeXt \cite{34} is a variant of ResNet \cite{39} with the building block shown in Fig. \ref{fig_resNet_architecture}. This block is very similar to the Inception module \cite{40}. They both comply with the split transform-merge paradigm. Our models are realized by the form of Fig. \ref{fig_resNet_architecture}. In the 3$\times$3 layer of the first block, downsampling of conv3, 4, and 5 is made by stride-2 convolutions in each stage, as suggested in \cite{39}. SGD is utilized with a mini-batch size of 256. The momentum is 0.9, and the weight decay is 0.0001. The initial value of learning rate is set to 0.1, and we divide it by a factor of 10 for three times following the schedule in \cite{39}. The weight initialization of \cite{39} is adopted, and we realize Batch Normalization (BN) \cite{41} right after the convolutions. ReLU is performed right after each BN.

\begin{table}[!t]
\renewcommand{\arraystretch}{1.3}
\caption{ELAPSED TIME (m: minutes, s: seconds) OF
TRAINING}
\label{table_costtime}
\centering
\begin{tabular}{|c|c|c|}
\hline
CNN           & LeNet-5  & ResNeXt \\ \hline
Elapsed Times & 1966m12s & 655m17s \\ \hline
\end{tabular}
\end{table}
By training the recognition model of each type of PEA in LeNet and ResNeXt, we aim to predict whether or not a type of PEA exists in an image/video patch. Note here we do not utilize a multi-target classification because of the non-exclusivity of PEAs ({\it i.e.} different types of PEAs coexist within one patch). Based on the above-mentioned two typical CNN networks, we individually train 6 types of PEA identification models. Let TP, FP, TN and FN denote the true positive, false positive, true negative, and false negative rates, respectively, the training and testing accuracy is defined as
$Accuracy = [TP/(TP+FP)+TN/(FN+TN)]/2$. Meanwhile, the cross-entropy loss function is adopted.

Table II lists the classification performance on our PEA datasets. From the results of each individual experimental data,  the recognition performance based on ResNeXt are significantly better than that solely based on LeNet. For example, in Table \ref{table_accuary}, the proposed blocking PEA recognition model yields a testing accuracy of 92.81\%, nearly 20\% higher than that of the LeNet ({\it i.e.} 71.62\%). Similar results are observed in the other PEA recognition models.
\begin{table}[!t]
\renewcommand{\arraystretch}{1.3}
\caption{Performance comparison of floating PEA recognition algorithm}
\label{table_comparission}
\centering
\begin{tabular}{|c|c|c|c|}
\hline
Algorithms               & Figure \ref{fig_An_example_of_texture_floating_detection} (b) & Figure \ref{fig_An_example_of_texture_floating_detection} (f)  & Image3000 \\ \hline
Ref{[}17{]}               & 96.1\%  & 54.92\%  & 65.17\%   \\ \hline
Proposed  & 95.85\% & 88.23\%  & 85.46\%   \\ \hline
\end{tabular}
\end{table}
Compared with LeNet, ResNeXt has more layers, and can learn more complex image high-dimensional features. By repeating a building block, ResNeXt is constructed. The building block aggregates a set of transformations with the same topology. Only a few hyper-parameters need to be set in a homogeneous and multi-branch architecture. Meanwhile, its bottleneck layer reduces the number of features. Thus the operation complexity of each layer reduces. Therefore, the computational complexity greatly reduces, while the speed and accuracy of the algorithm improves.
\begin{figure*}[!t]
  \centering
  \subfigure[]{
    \includegraphics[width=1.5in,height=3.5cm]{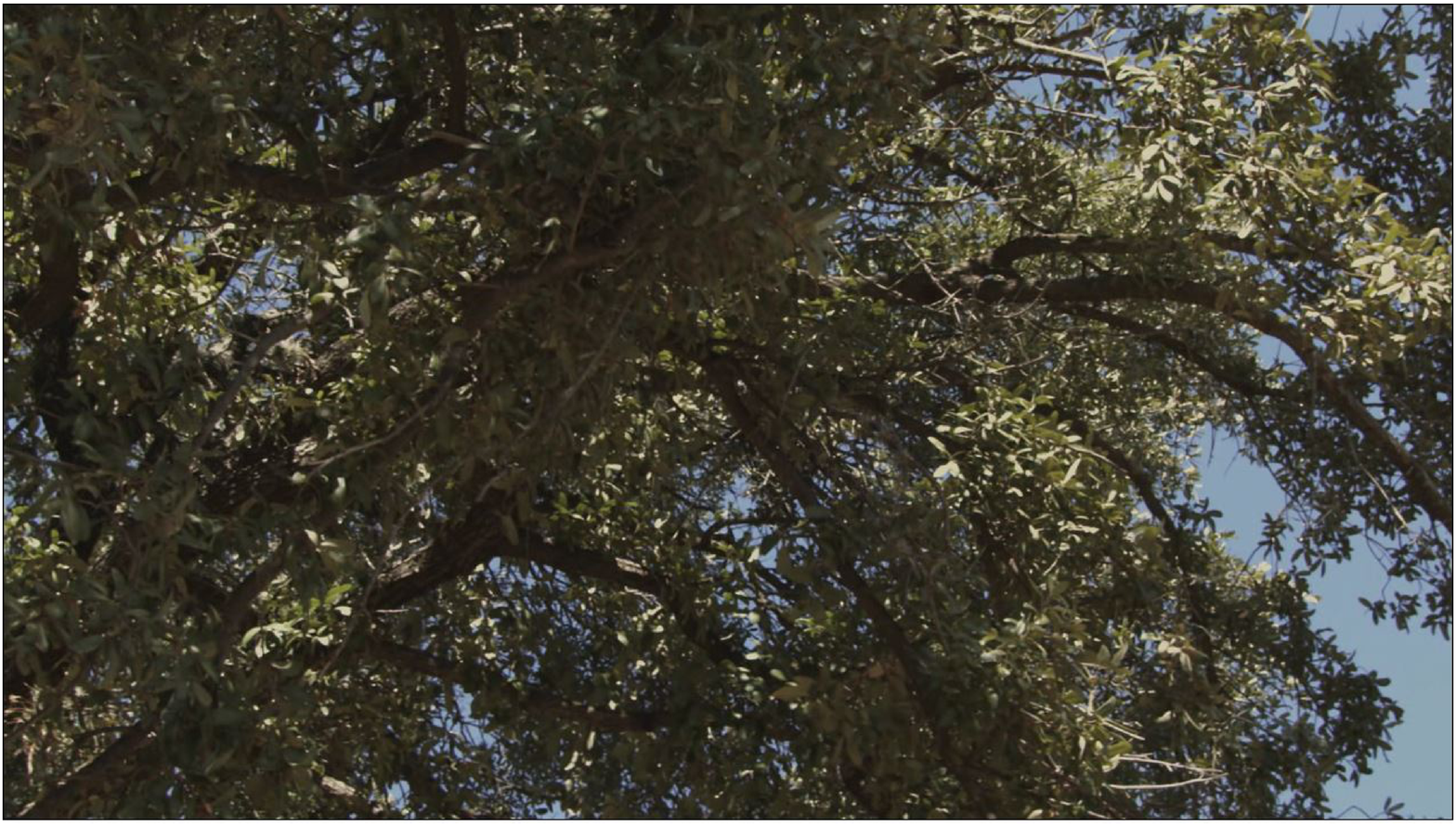}
  }
  \subfigure[]{
    \includegraphics[width=1.5in,height=3.5cm]{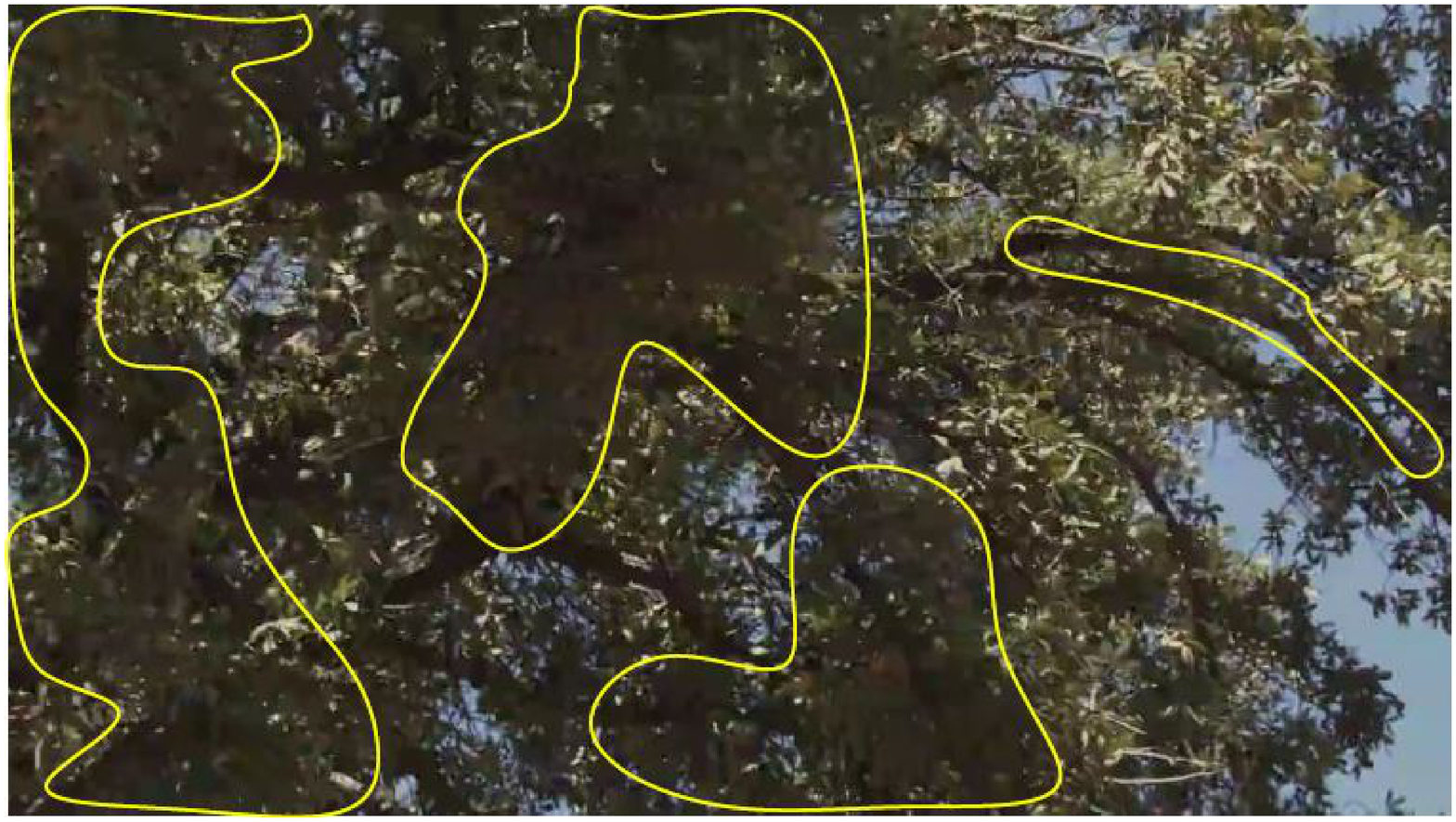}
  }
  \subfigure[]{
    \includegraphics[width=1.5in,height=3.5cm]{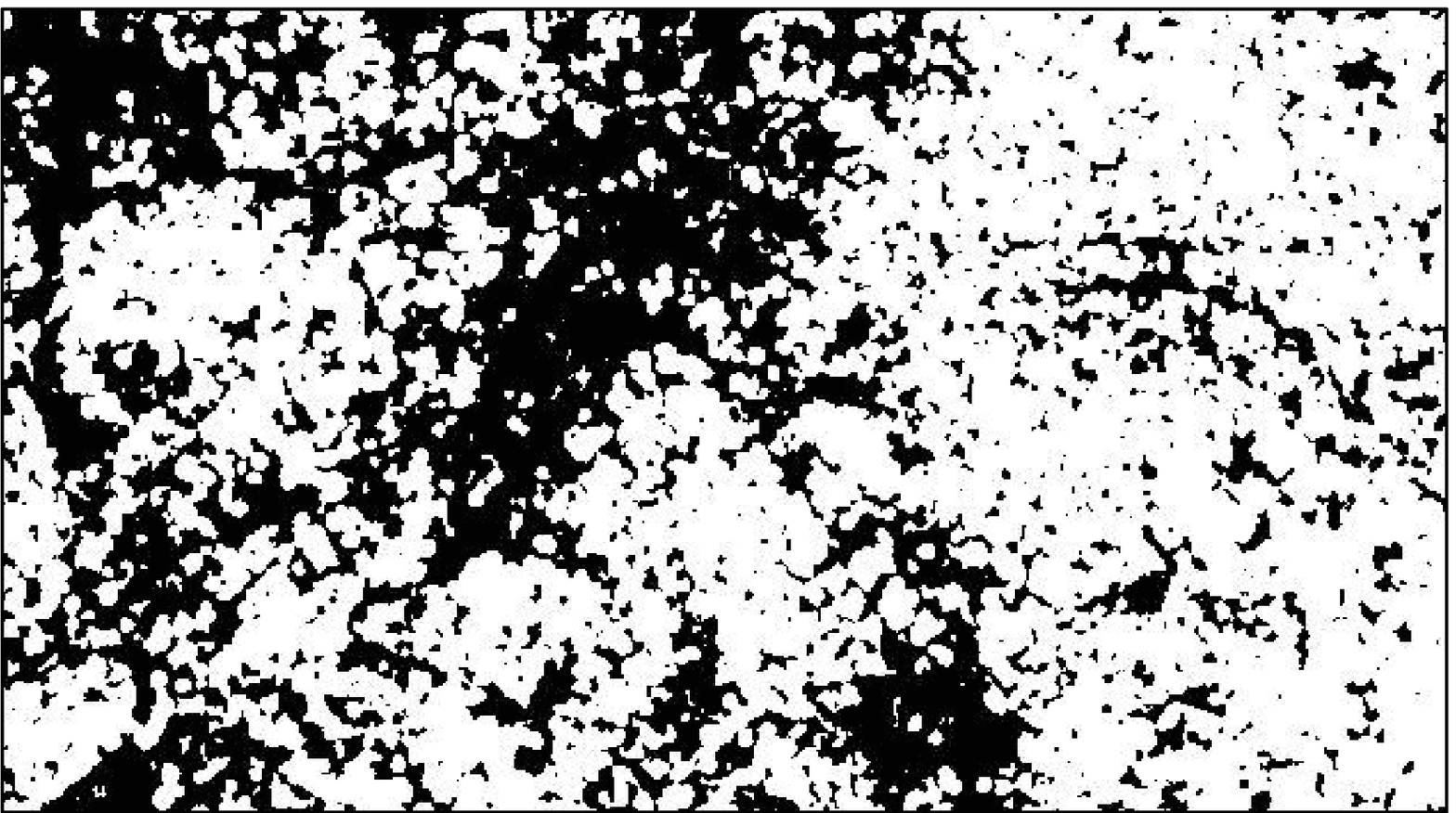}
  }
   \subfigure[]{
    \includegraphics[width=1.5in,height=3.5cm]{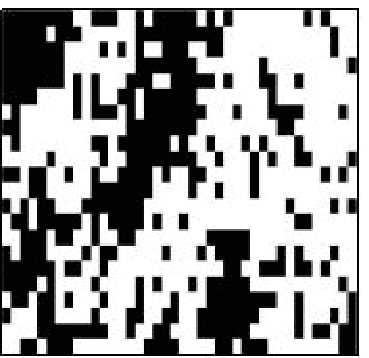}
  }
  \subfigure[]{
    \includegraphics[width=1.5in,height=3.5cm]{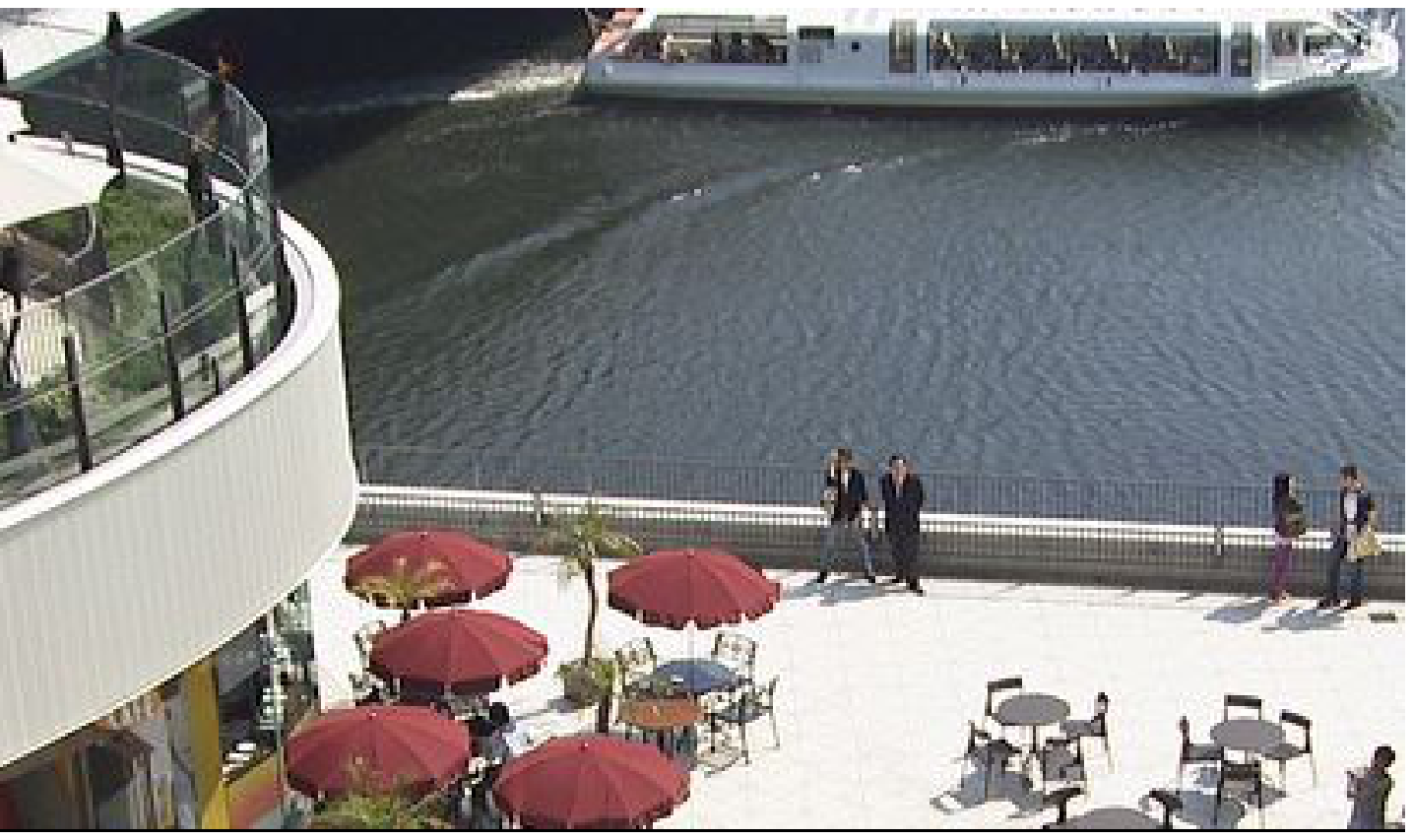}
  }
   \subfigure[]{
    \includegraphics[width=1.5in,height=3.5cm]{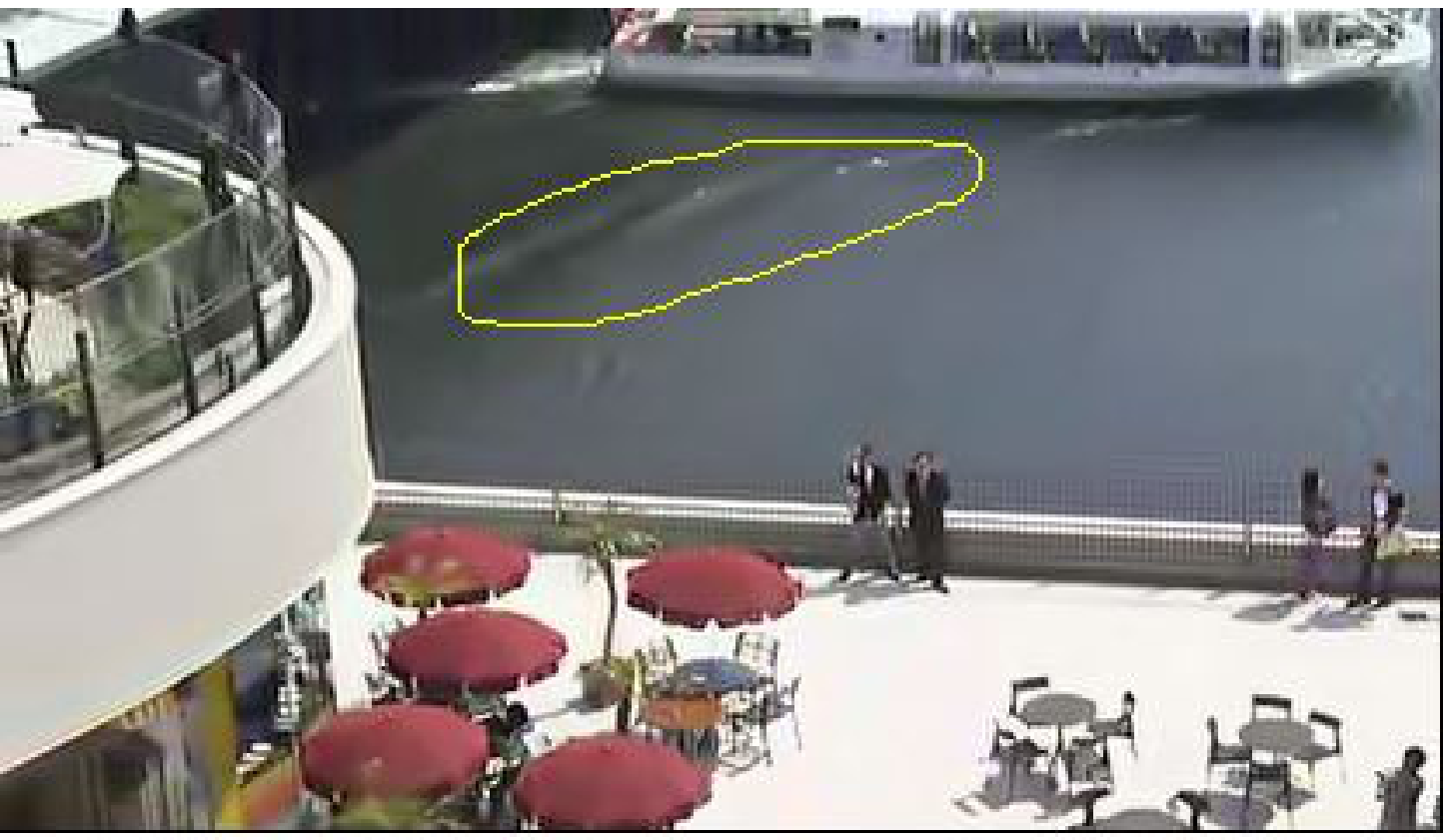}
  }
   \subfigure[]{
    \includegraphics[width=1.5in,height=3.5cm]{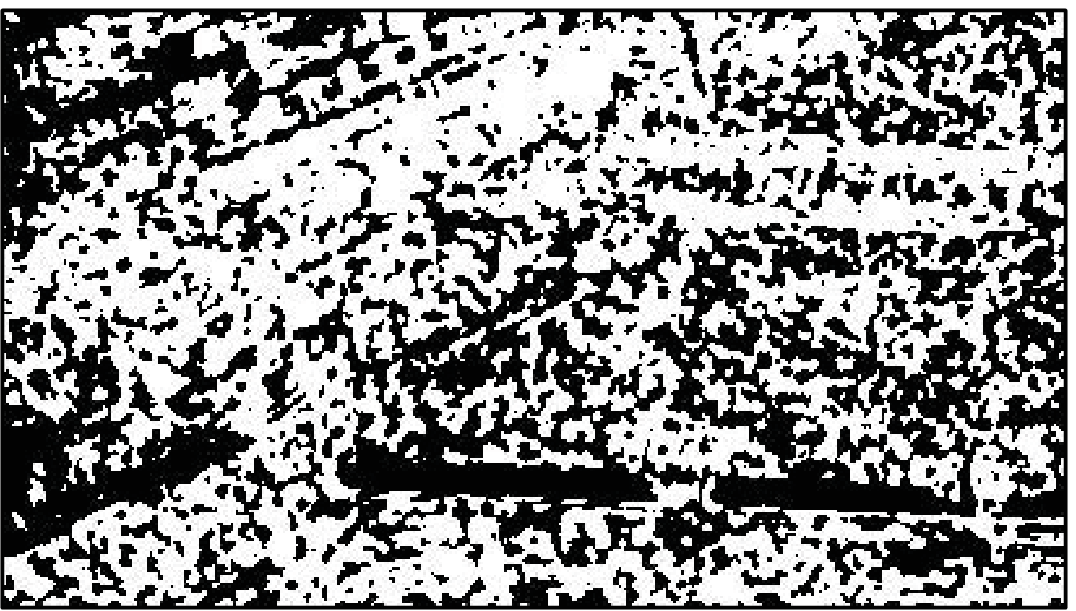}
  }
   \subfigure[]{
    \includegraphics[width=1.5in,height=3.5cm]{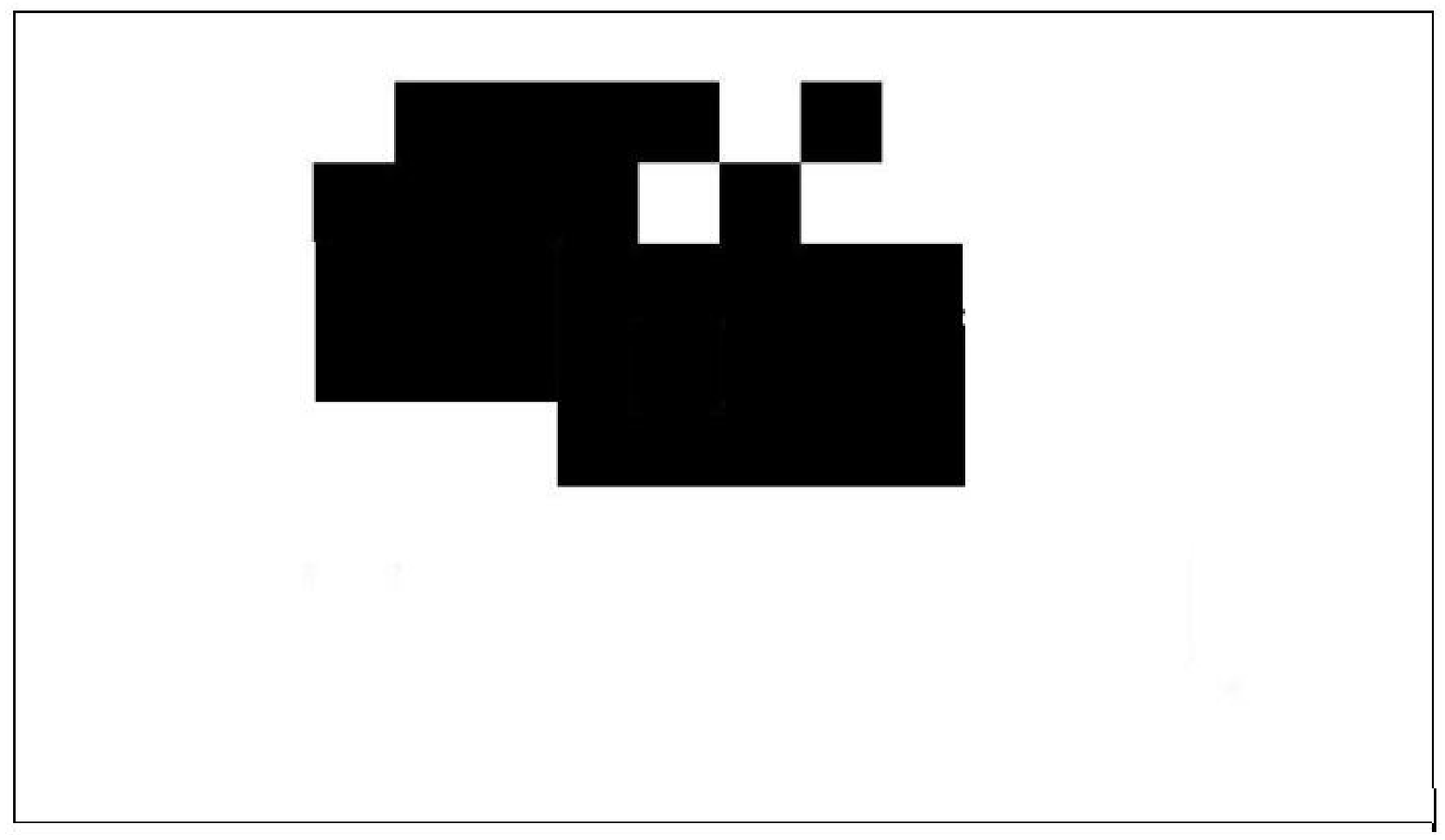}
  }
\caption{An example of floating PEA detection.}
\label{fig_An_example_of_texture_floating_detection}
\end{figure*}
The computational complexity of the training and testing procedures using the LeNet and ResNeXt is summarized in Table \ref{table_costtime}. ResNeXt is much faster than LeNet because of the bottleneck layer. The training process requires a large number of iterations and is relatively time-consuming.

\begin{figure*}[!ht]
  \centering
  \subfigure[PEA pattern with spatial PEA(s)]{
    \includegraphics[width=3.2in,height=4cm]{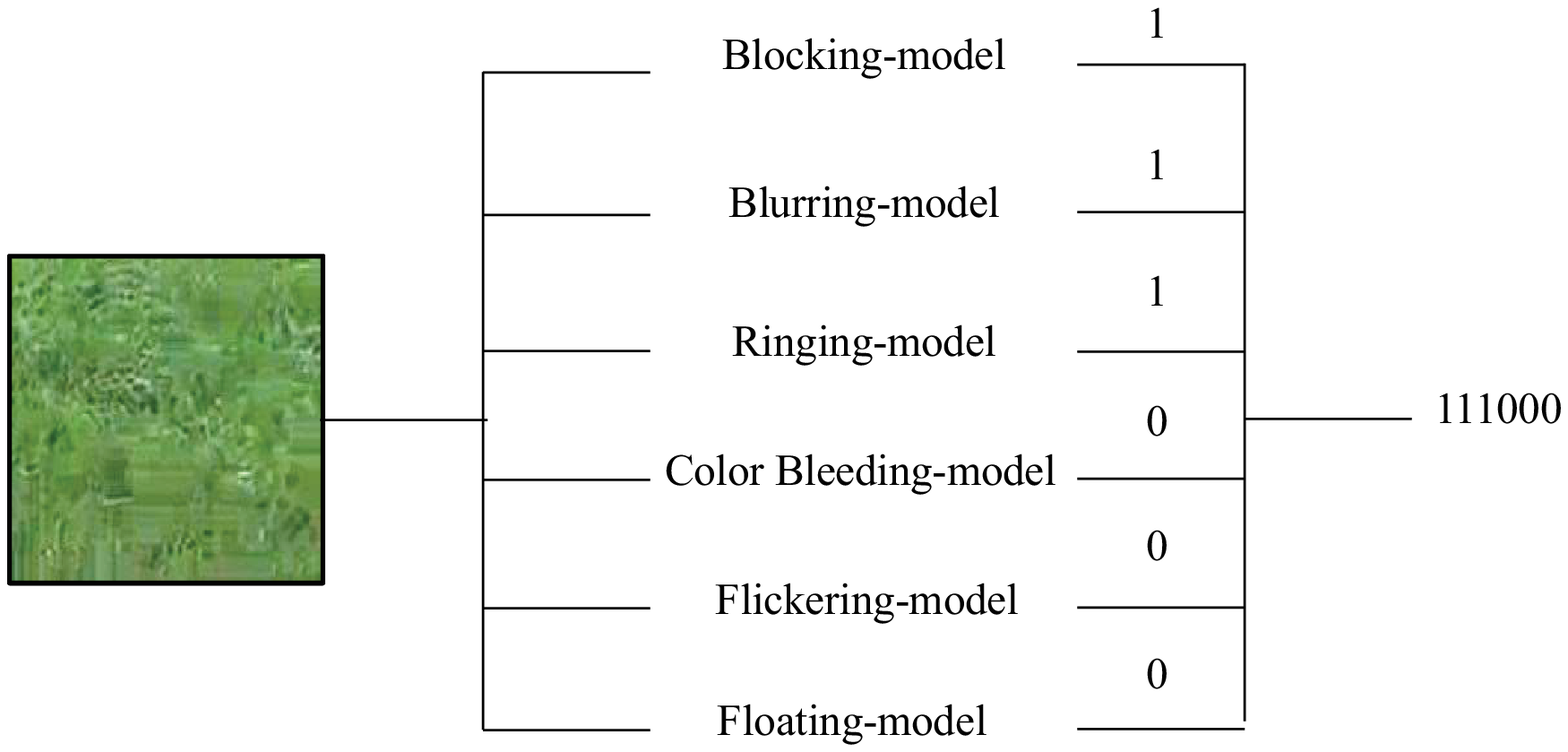}
  }
  \subfigure[PEA pattern with temporal PEA(s)]{
    \includegraphics[width=3.2in,height=4cm]{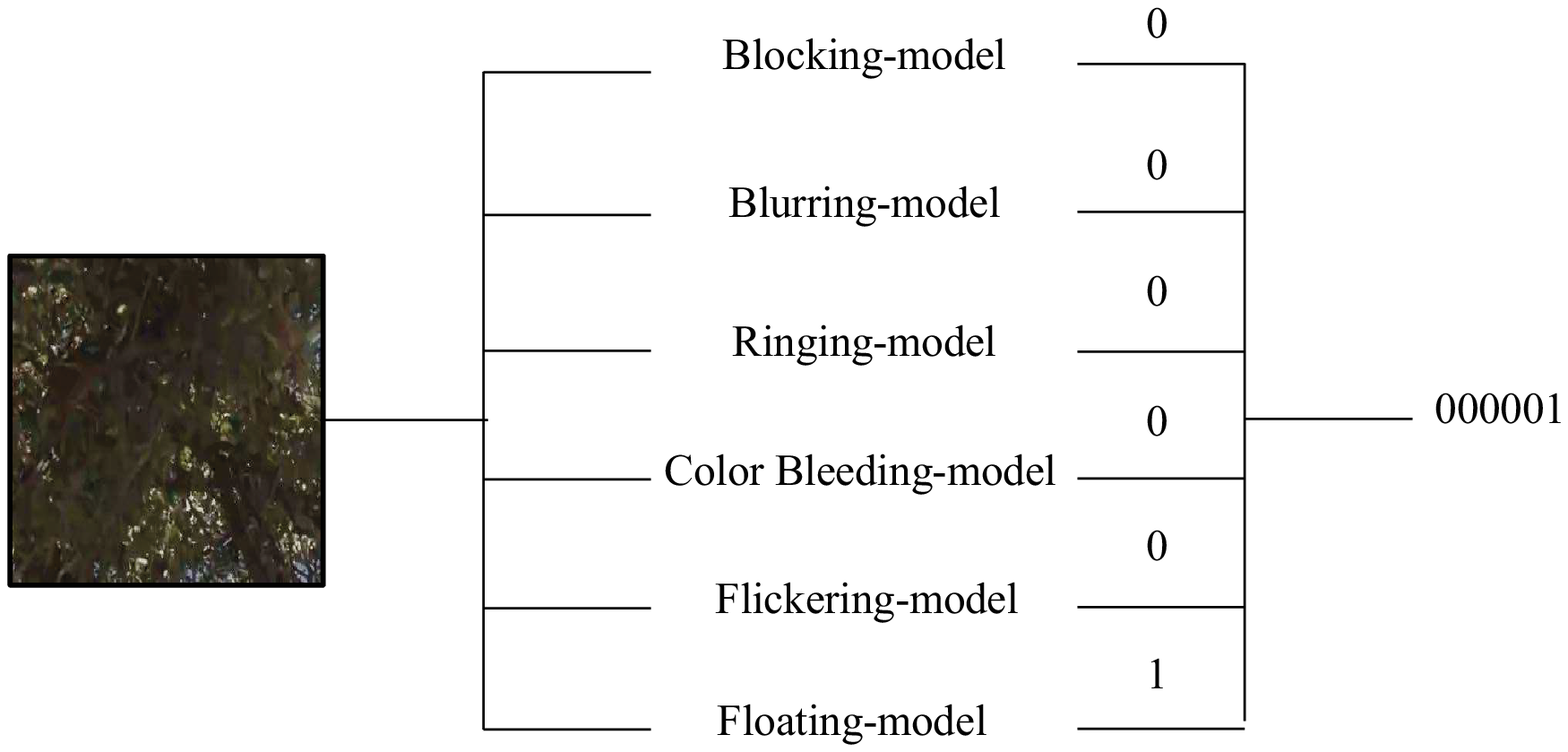}
  }
\caption{The PEA pattern of image patches.}
\label{fig_The_PEA_detection}
\end{figure*}

\subsection{Comparison with other benchmarks}
In order to better illustrate the advantages of the proposed recognition, we compare it with the floating PEA detection method in \cite{17}, in which the low-level coding features were extracted to estimate the spatial distribution of floating. Fig. \ref{fig_An_example_of_texture_floating_detection} (a) and (e) are two original frames, respectively, and Fig. \ref{fig_An_example_of_texture_floating_detection} (b) and (f) are their compressed frames, coded by HEVC with Qp = 42, where the visual floating regions are marked manually. Fig. \ref{fig_An_example_of_texture_floating_detection} (c) is the floating map generated by \cite{17}, where black regions indicate the floating artifacts. Fig. \ref{fig_An_example_of_texture_floating_detection} (d) is the result of the proposed PEA recognition model. In this case, both methods performs reasonably well in floating detection. However, the algorithm in \cite{17} requires content-dependent parameter adjustment and does not generalize consistently. For example, Fig. \ref{fig_An_example_of_texture_floating_detection} (g) fails to detect the actual floating region. Compared Fig. \ref{fig_An_example_of_texture_floating_detection} (g) with Fig. \ref{fig_An_example_of_texture_floating_detection} (h), the proposed floating PEA recognition algorithm performs clearly better. The floating detection accuracy is given in Table \ref{table_comparission}. In addition, we randomly select 3000 test images, and the performance comparison results are illustrated in the last column of the Table \ref{table_comparission}. It appears that the proposed floating PEA recognition model consistently outperforms \cite{17}.
\begin{figure*}[!ht]
  \centering
  \subfigure[A compressed frame]{
    \includegraphics[width=1.5in,height=3.5cm]{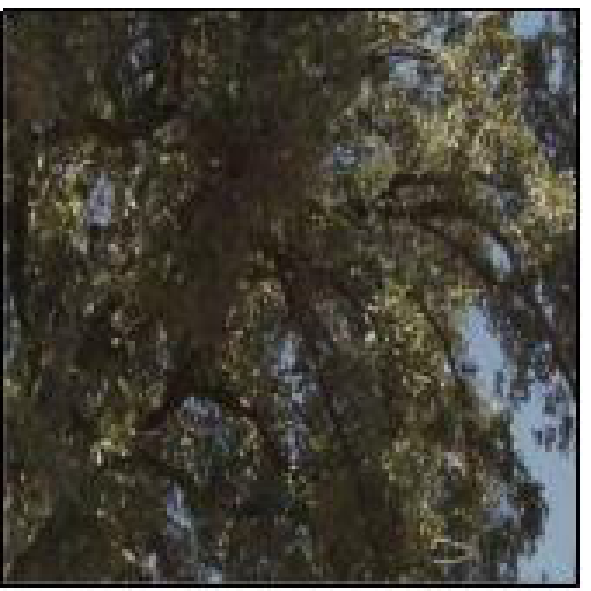}
  }
  \subfigure[Blocking artifact]{
    \includegraphics[width=1.5in,height=3.5cm]{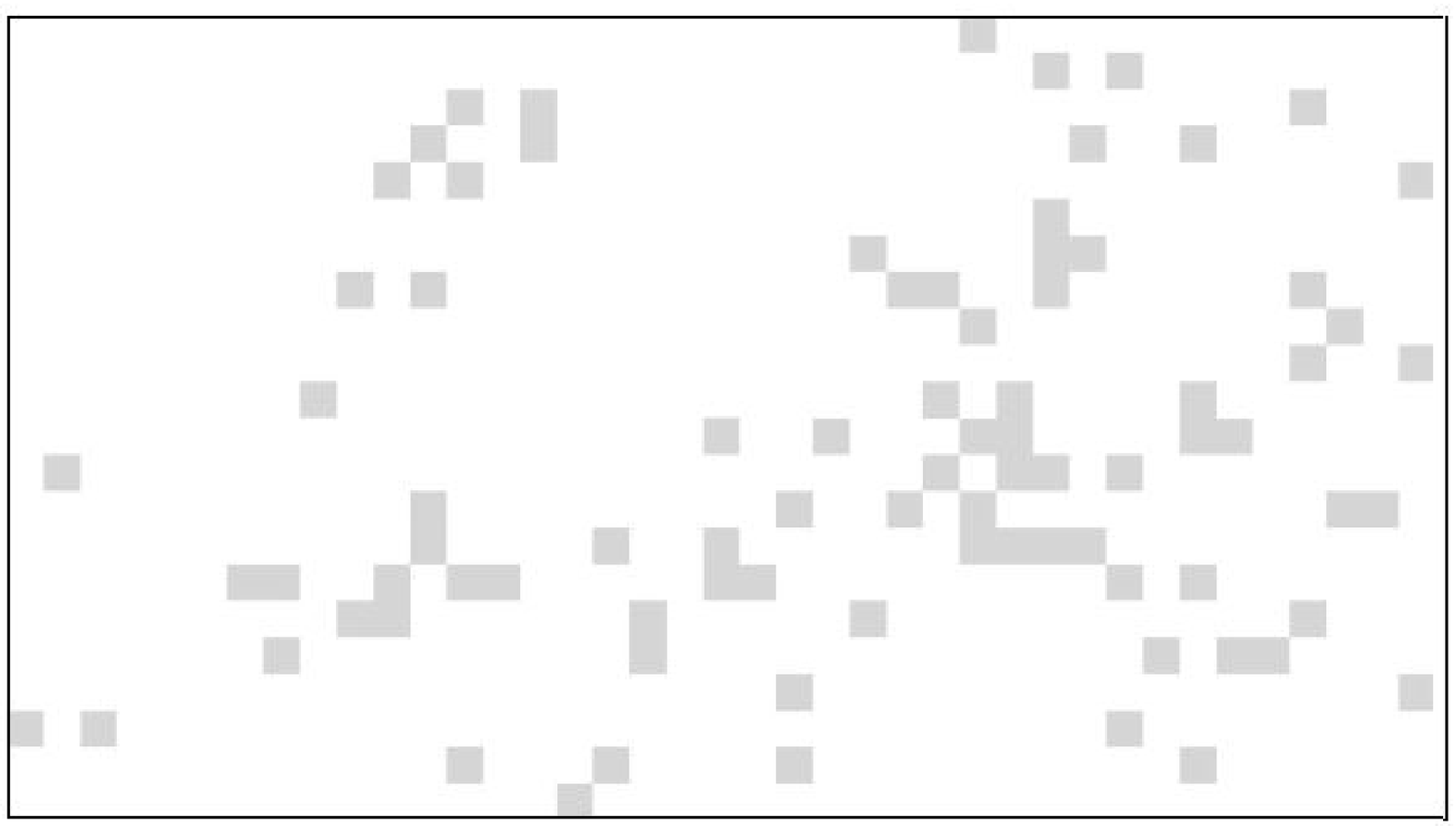}
  }
  \subfigure[Blurring artifact]{
    \includegraphics[width=1.5in,height=3.5cm]{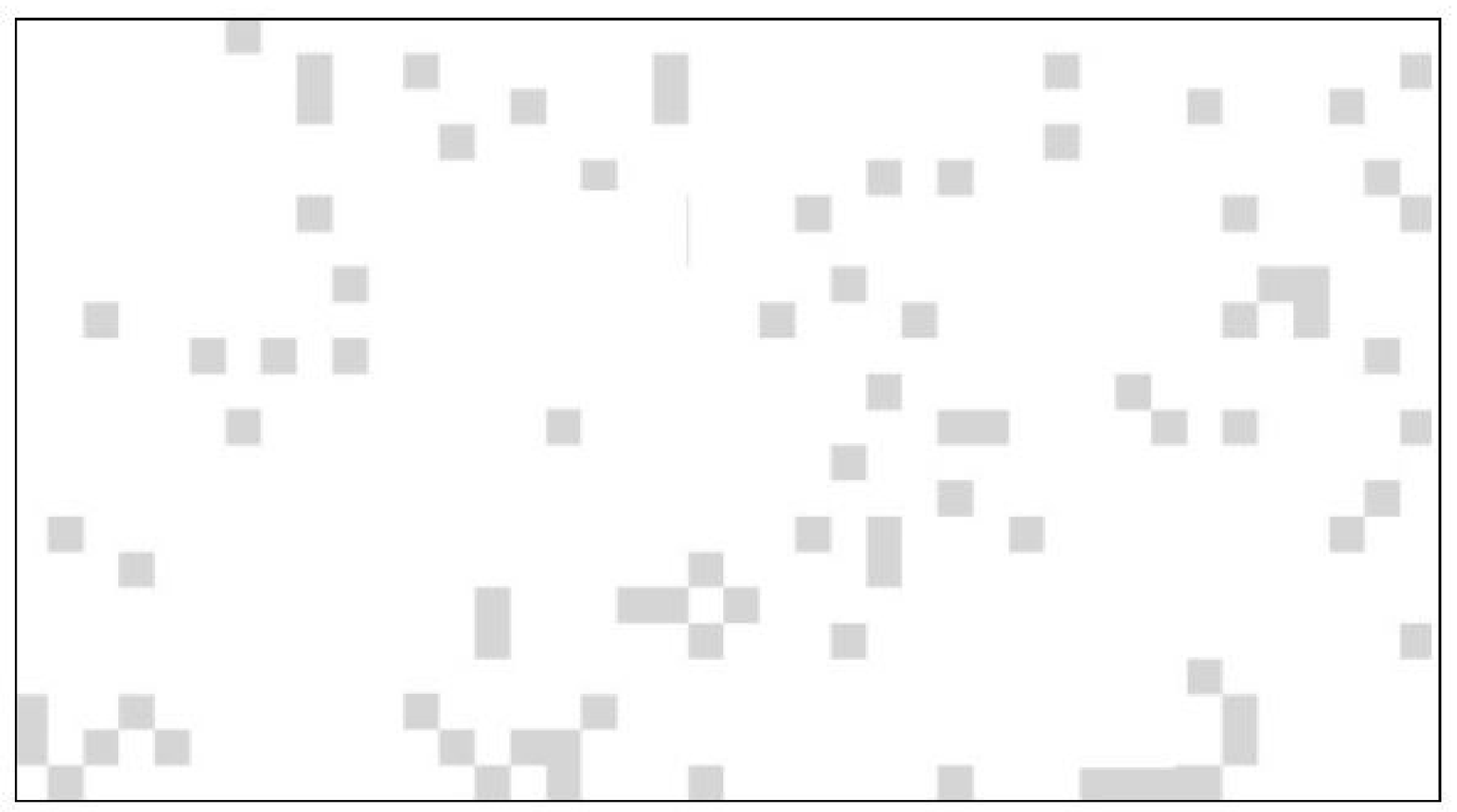}
  }
   \subfigure[Ringing artifact]{
    \includegraphics[width=1.5in,height=3.5cm]{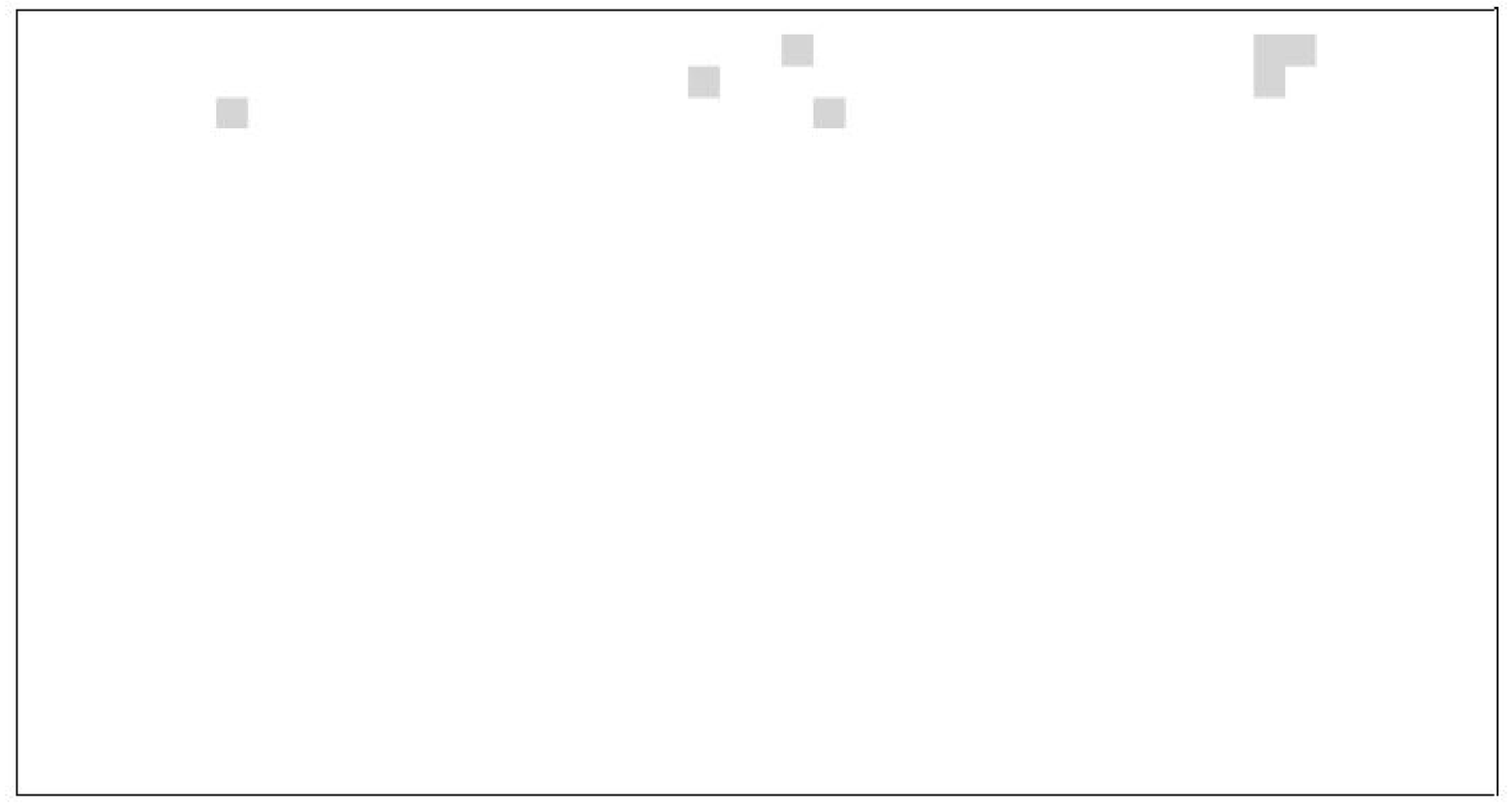}
  }
  \subfigure[Color bleeding artifact]{
    \includegraphics[width=1.5in,height=3.5cm]{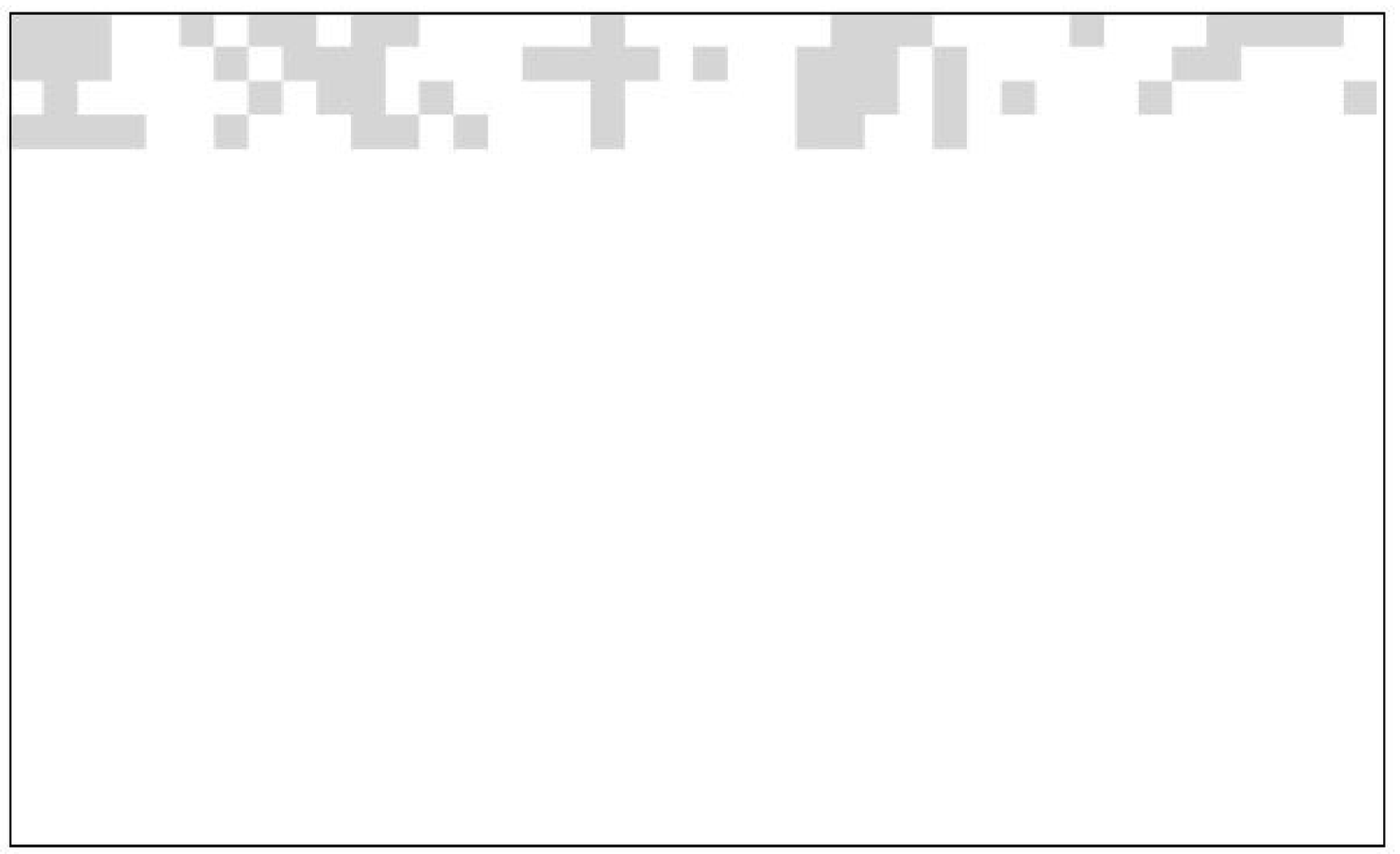}
  }
   \subfigure[Flickering artifact]{
    \includegraphics[width=1.5in,height=3.5cm]{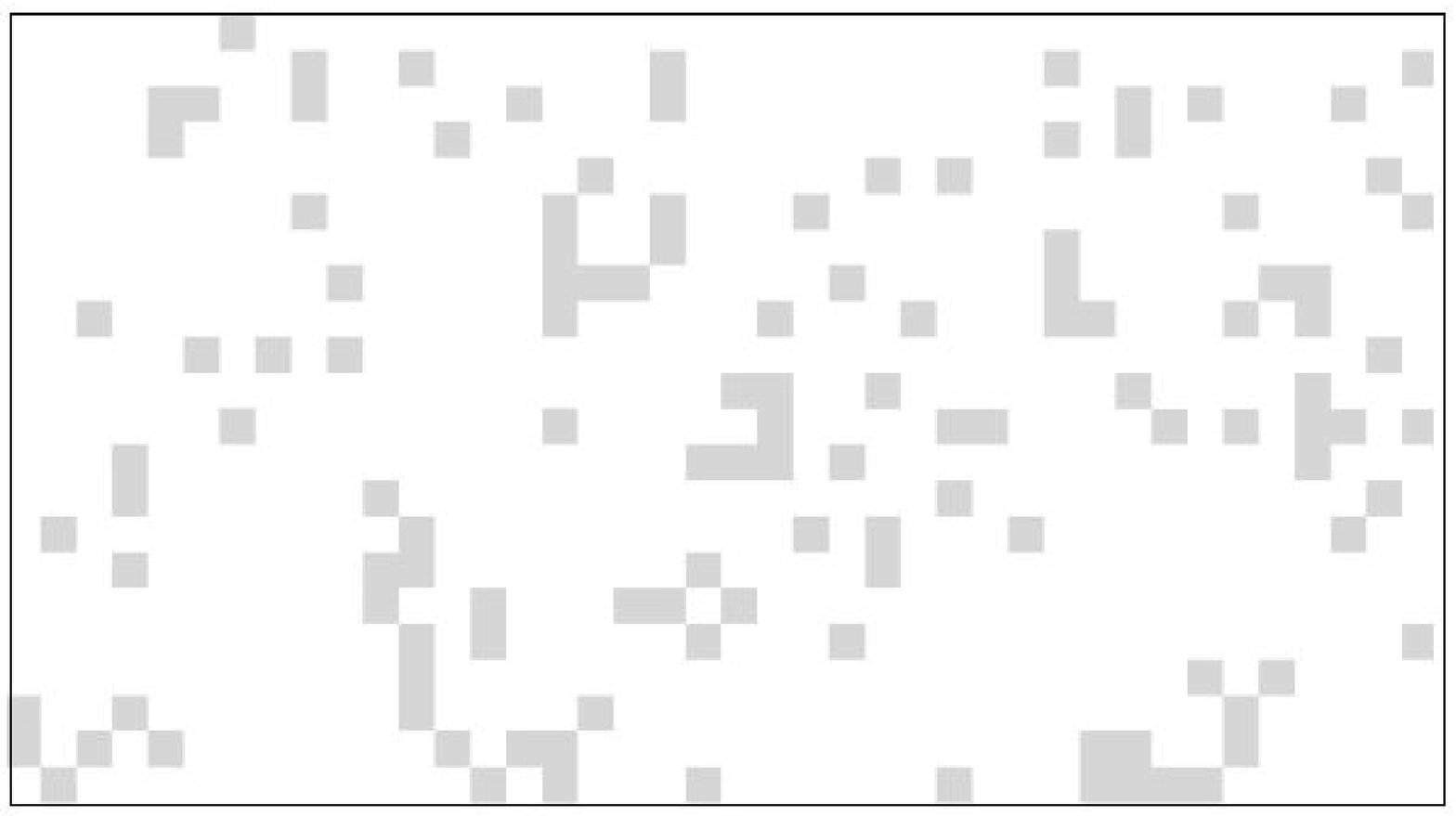}
  }
   \subfigure[Floating artifact]{
    \includegraphics[width=1.5in,height=3.5cm]{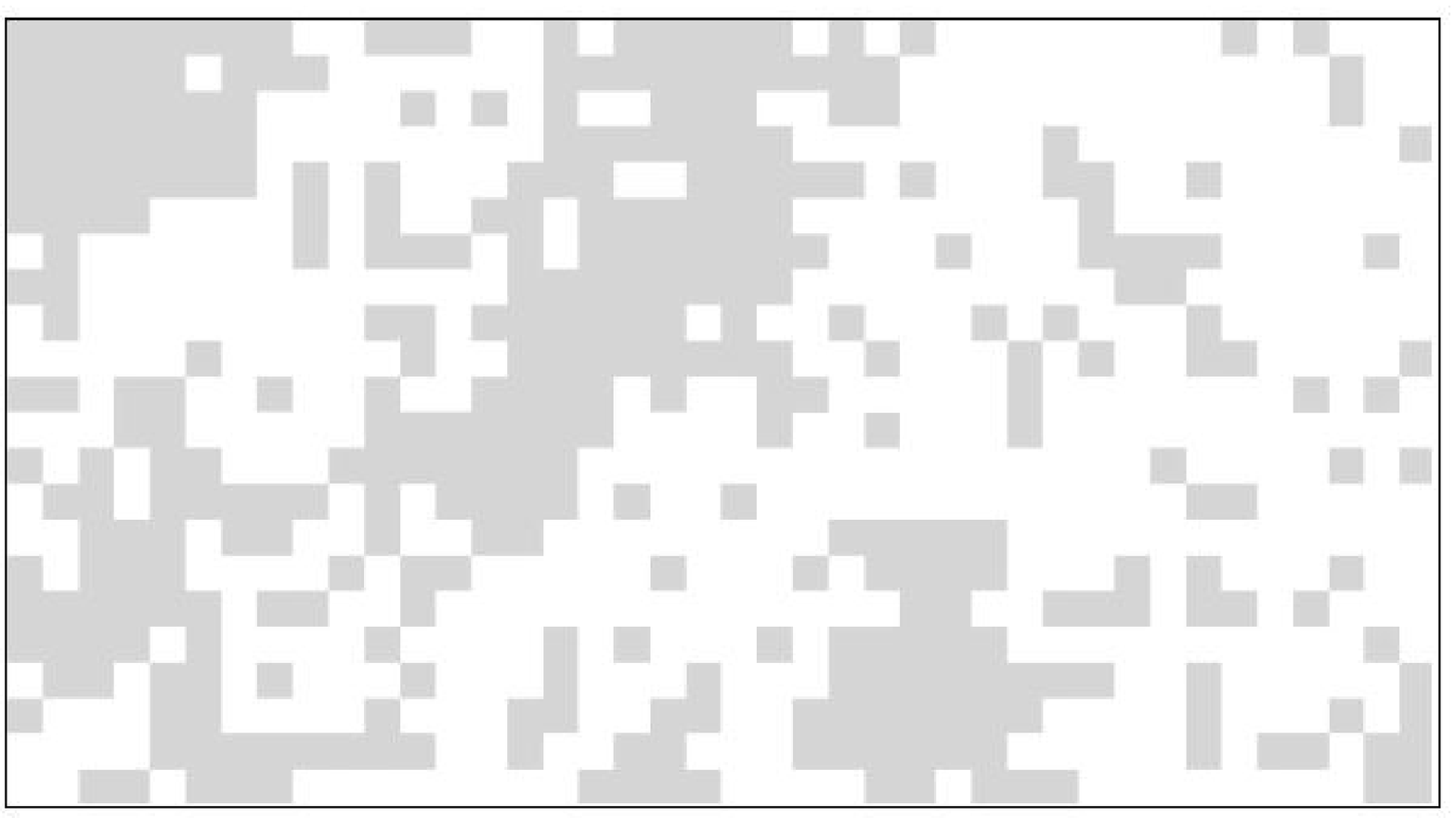}
  }
   \subfigure[Combined artifacts]{
    \includegraphics[width=1.5in,height=3.5cm]{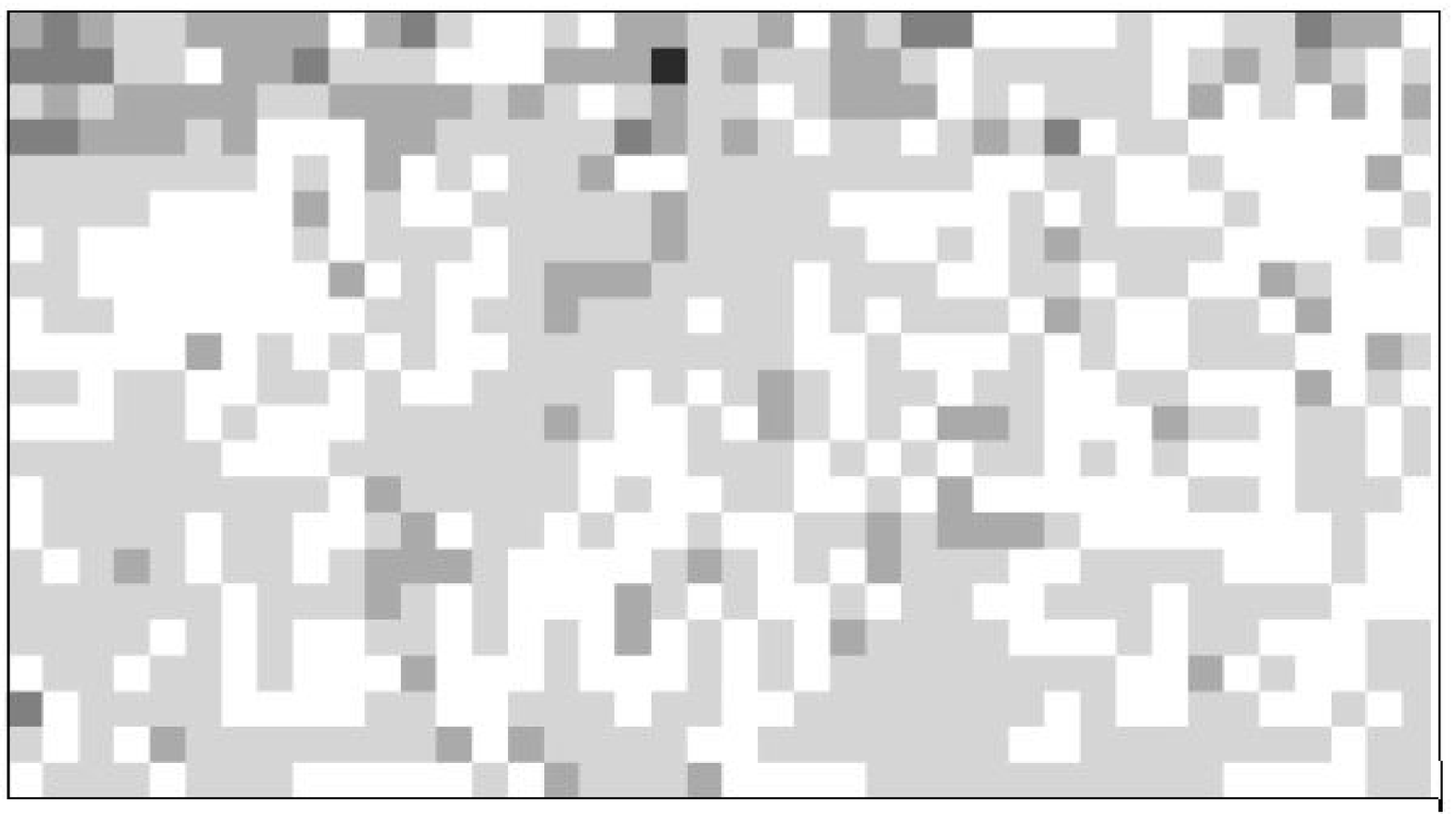}
  }
\caption{ The individual and overall PEA distributions of a frame.}
\label{fig_PEA_intensity_frame}
\end{figure*}

\subsection{The overall PEA intensity}

By combining the 6 PEA recognition models, we obtain two hybrid PEA metrics: a local PEA metric, namely PEA pattern, and a holistic PEA metric, namely PEA intensity. A PEA pattern is represented as a 6-bin value, each contains a binary value representing the existence of blurring, blocking, ringing, color bleeding, flickering and floating artifacts respectively. We set it 1 if its corresponding PEA exists; otherwise 0. To intuitively show the PEA pattern, we present two examples in Fig. \ref{fig_The_PEA_detection}. In Fig. \ref{fig_The_PEA_detection} (a), blurring, blocking and ringing artifacts exist in this patch, thus its PEA pattern is labeled as 111000; In Fig. \ref{fig_The_PEA_detection} (b), floating artifact exists in this patch, thus its PEA pattern is labeled as 000001. This pattern denotes the feature vector of a video patch in terms of PEAs and thus can be further utilized in vision-based video processing. In addition, we summarize the distributions of all types of PEAs in Fig. \ref{fig_PEA_intensity_frame}. It is observed that for a video frame, the distributions of PEAs differ from each other in which all types of PEAs may not be observed simultaneously. Therefore, only a combination of PEAs, such as in Fig. \ref{fig_PEA_intensity_frame} (h), shows the impacts of PEAs on visual quality. We introduce a new metric, PEA intensity, as the percentage of positive binaries ({\it i.e.} value 1) within a patch, to illustrate this overall impact. The PEA patterns, 111000 and 000111, have the same PEA intensity because there are 3 positive binaries in both patterns.

\begin{figure*}[!t]
 \centering
 \subfigure[Blocking intensity]{
    \includegraphics[width=3.4in,height=5.6cm]{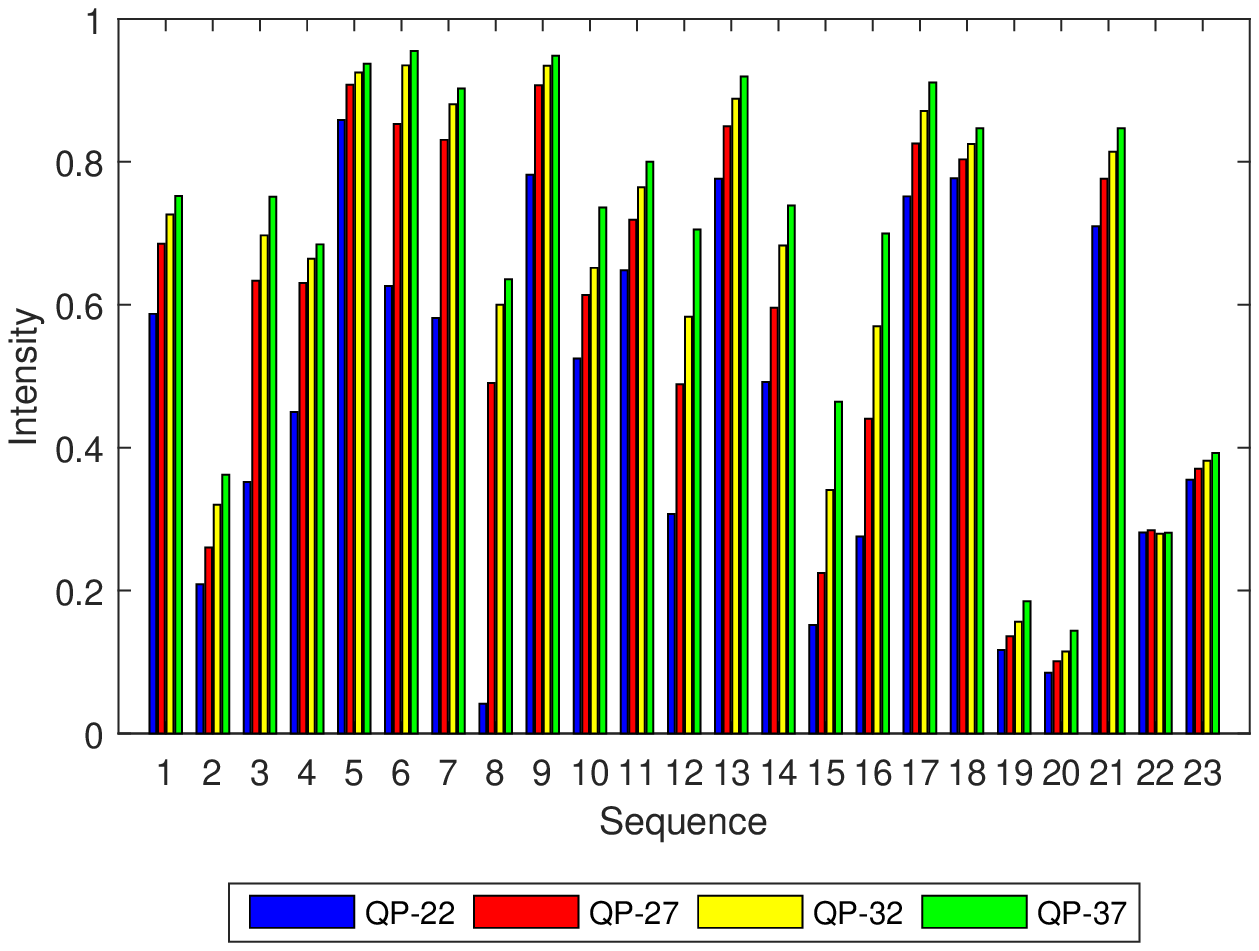}
  }
   \subfigure[Blurring intensity]{
    \includegraphics[width=3.4in,height=5.6cm]{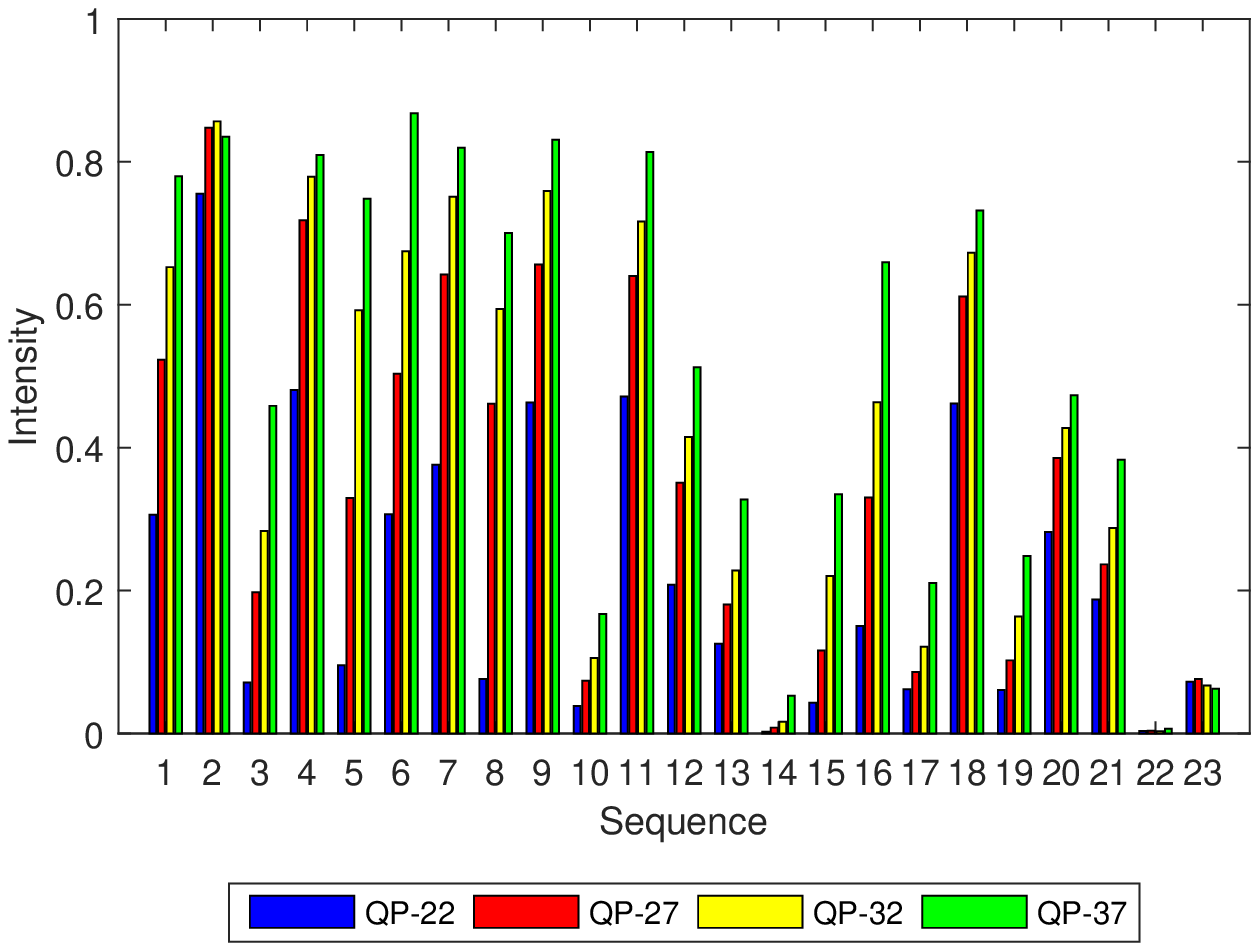}
  }
 \subfigure[ Color Bleeding intensity]{
    \includegraphics[width=3.4in,height=5.6cm]{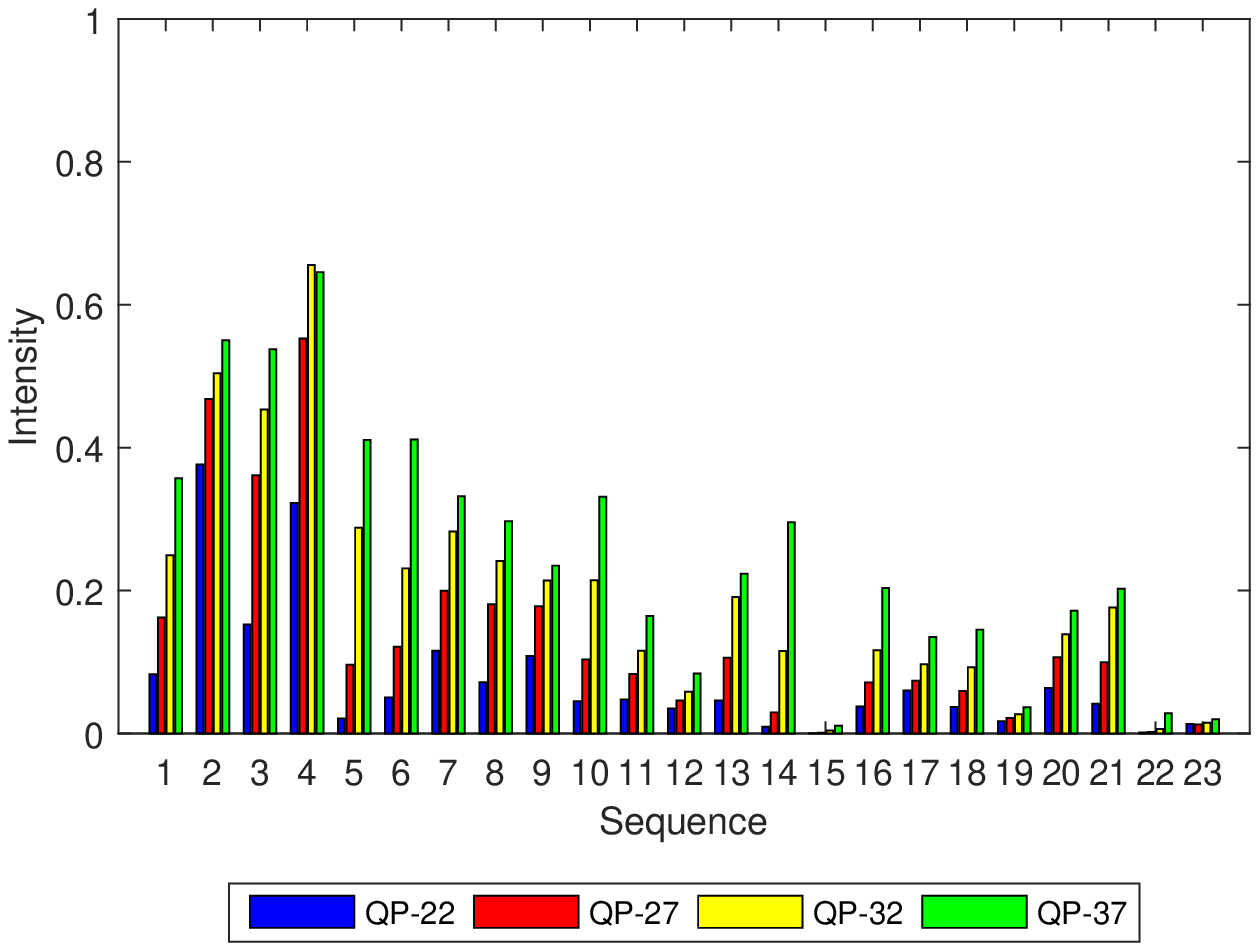}
  }
   \subfigure[Ringing intensity]{
    \includegraphics[width=3.4in,height=5.6cm]{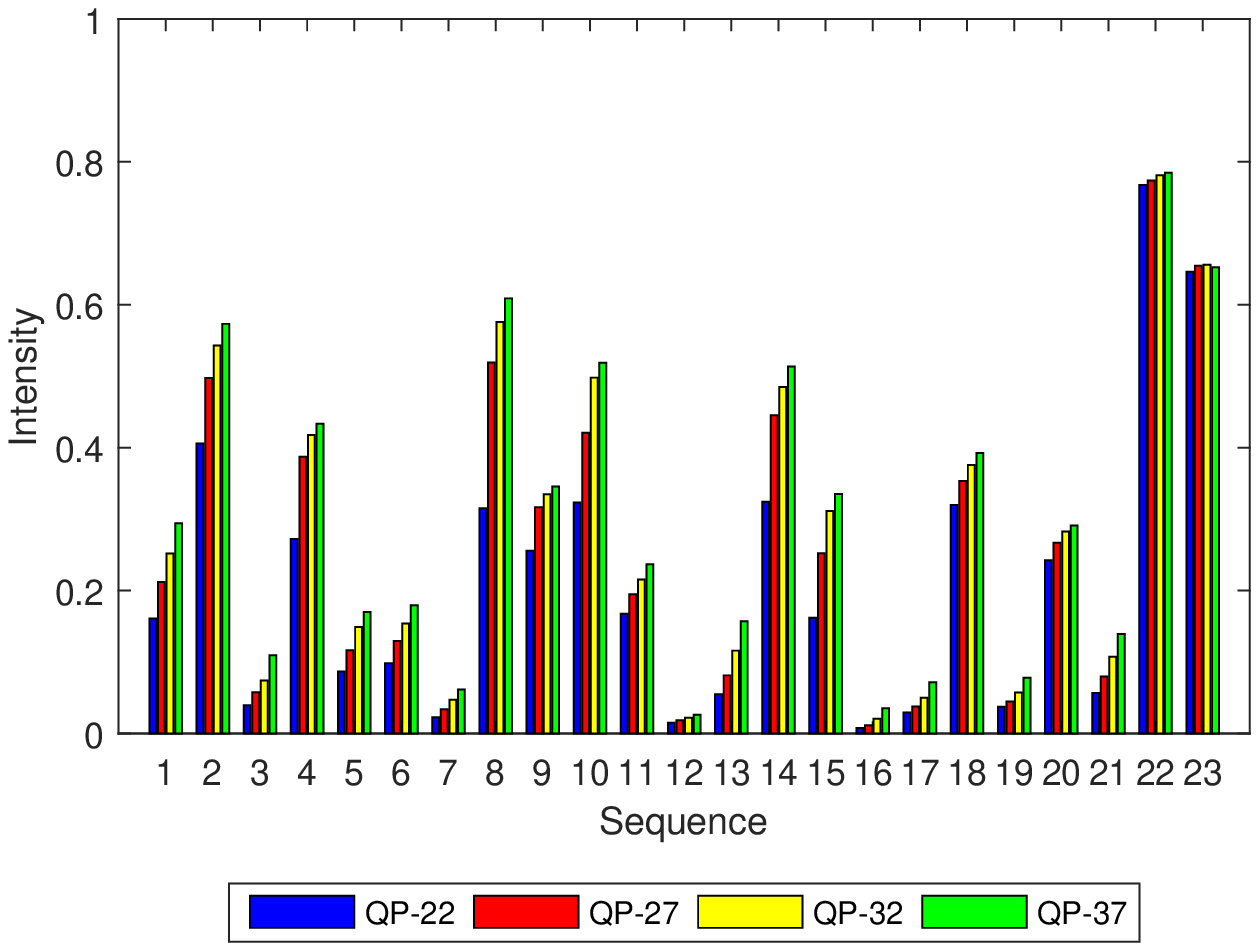}
  }
  \subfigure[Flickering intensity]{
    \includegraphics[width=3.4in,height=5.6cm]{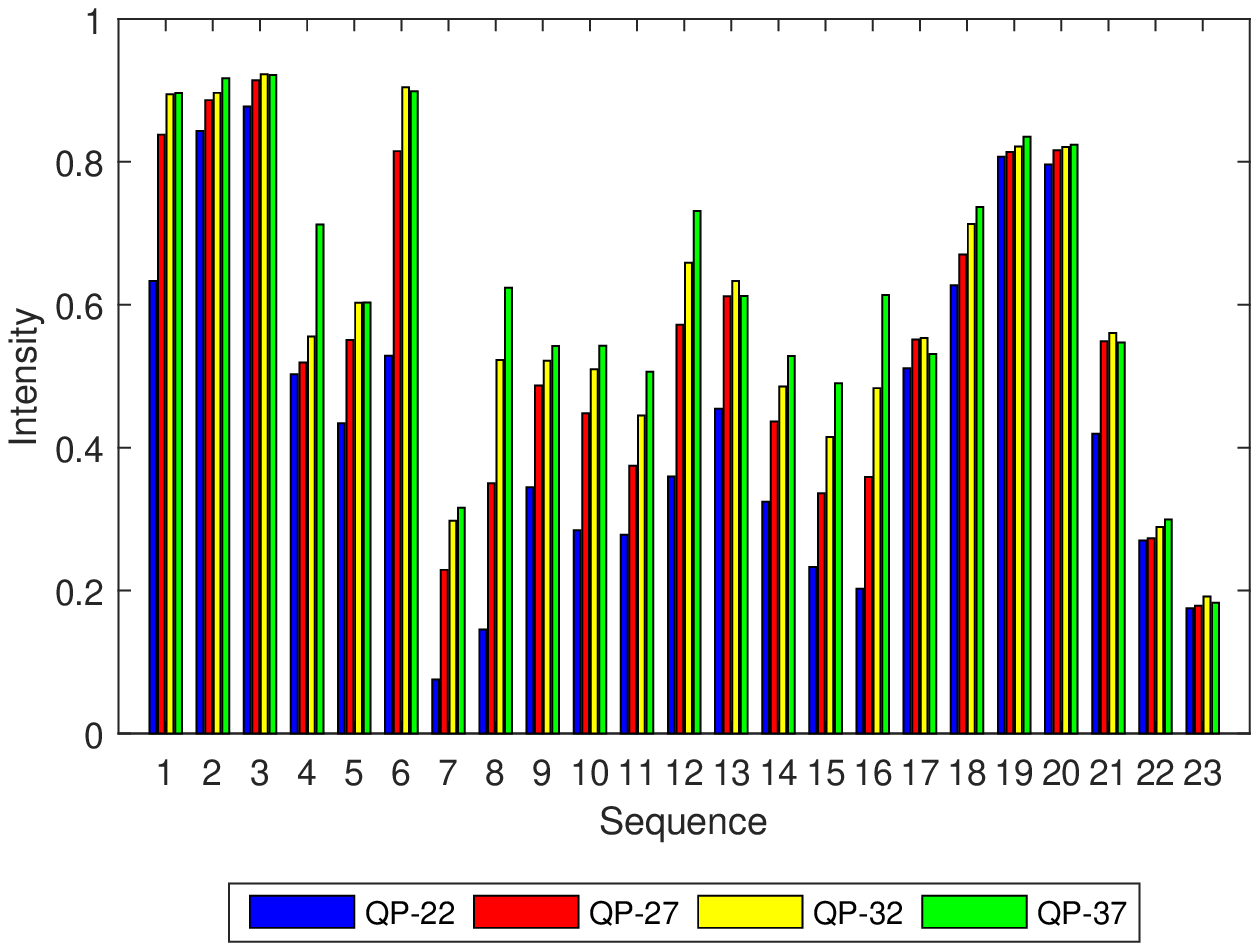}
  }
   \subfigure[Floating intensity]{
    \includegraphics[width=3.4in,height=5.6cm]{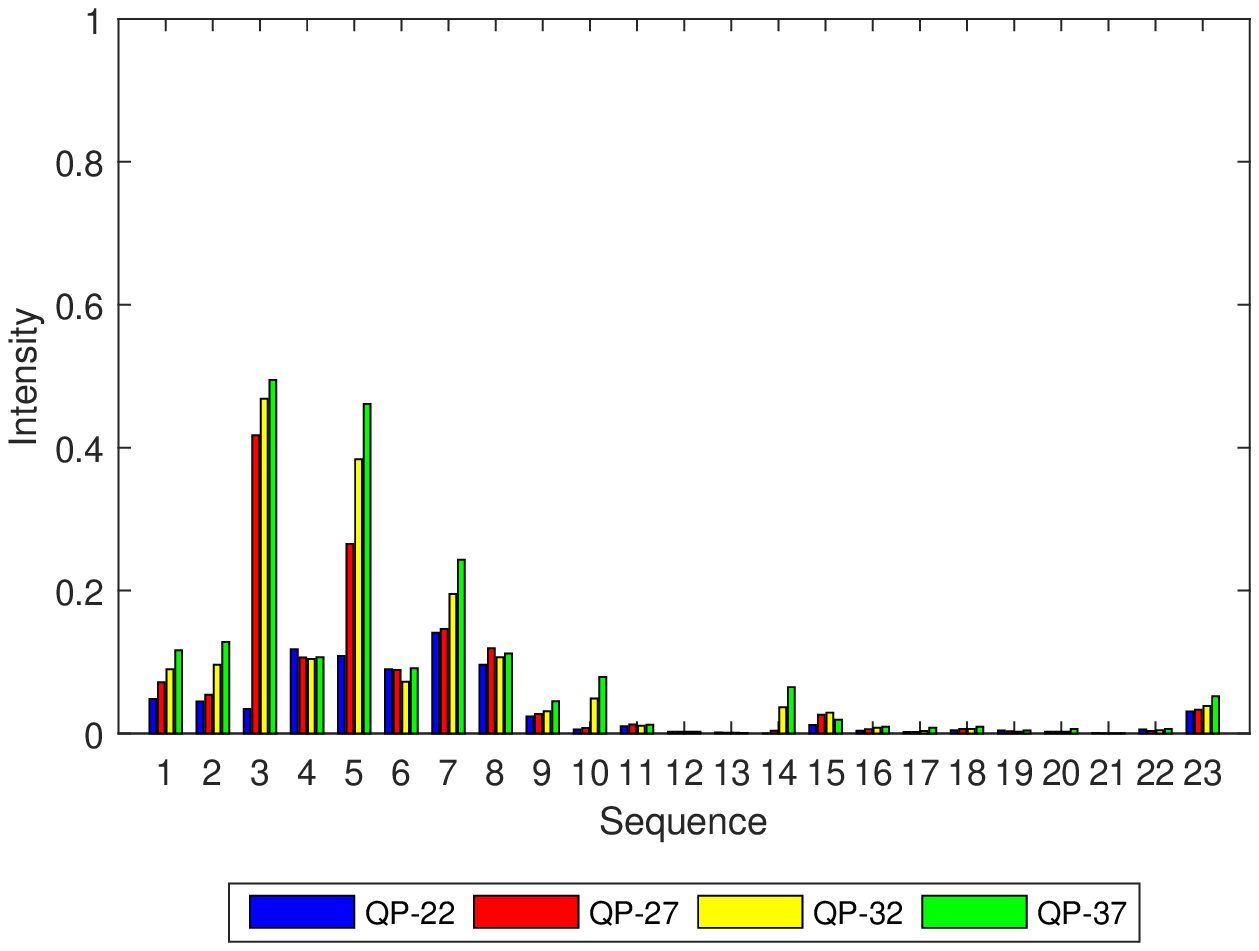}
  }

  \caption{The individual PEA intensity for each type of PEA.}
  \label{fig_PEA1_detection}
\end{figure*}

For a video sequence, its PEA intensity is then defined as the average PEA intensity of all non-overlapping patches. In Fig. \ref{fig_PEA1_detection}, the overall PEA intensities of all CTC sequences are measured and presented. Several conclusions can be drawn here. Firstly, the overall PEA intensity is, in general, positively correlated to the Qp value. For almost all types of PEAs and videos, the PEA intensity grows with a higher Qp. This fact highlights the importance of quantization and information loss in the generation mechanism of PEAs. As discussed before, the potential origin of spatial artifacts are interpreted as the loss of high frequency signals, chrominance signals and inconsistency of information loss between boundaries, while the temporal artifacts are possibly produced by inconsistent information loss between frames. Therefore, the fact that Qp influences PEA intensity is compatible with the above interpretations and also provides guidance to detailed explorations on the generation mechanism of PEAs.

Secondly, the PEA intensity is content-dependent, as it varies subject to video contents. For example, the sequences {\em SlideEditing} (1280$\times$720, No.22), {\em SlideShow} (1280$\times$720, No.23) have lower PEA intensities in terms of blocking, blurring and floating; on the other hand, more color bleeding, ringing and flickering artifacts are identified. The sequence {\em Kimono} (1920$\times$1080, No.5) has severe intensities for almost all types of PEAs while the sequence  {\em BQSquare} (416$\times$240, No.15) is with low intensities for almost all PEAs. This implies that the video characteristics, including texture and motion, may have an impact on the PEA intensity when being compressed. It may also provide useful instructions for content-aware video coding optimization.

Thirdly, the frequencies of PEAs can be different subject to its type. In this database, the intensities of blocking, color bleeding and flickering are significant compared with other PEAs including blurring, ringing and floating. Furthermore, the impact on visual quality changes for different types of PEAs. All types of PEAs may not have the same impact on HVS and the visual quality of users may be dominated by parts of PEAs, as concluded in \cite{18}. We put this in future work to explore how PEA detections should be combined to best evaluate their impact on visual quality.

\begin{figure*}[!t]
 \centering
 \subfigure[Spatial PEA intensity]{
    \includegraphics[width=3.4in,height=6.5cm]{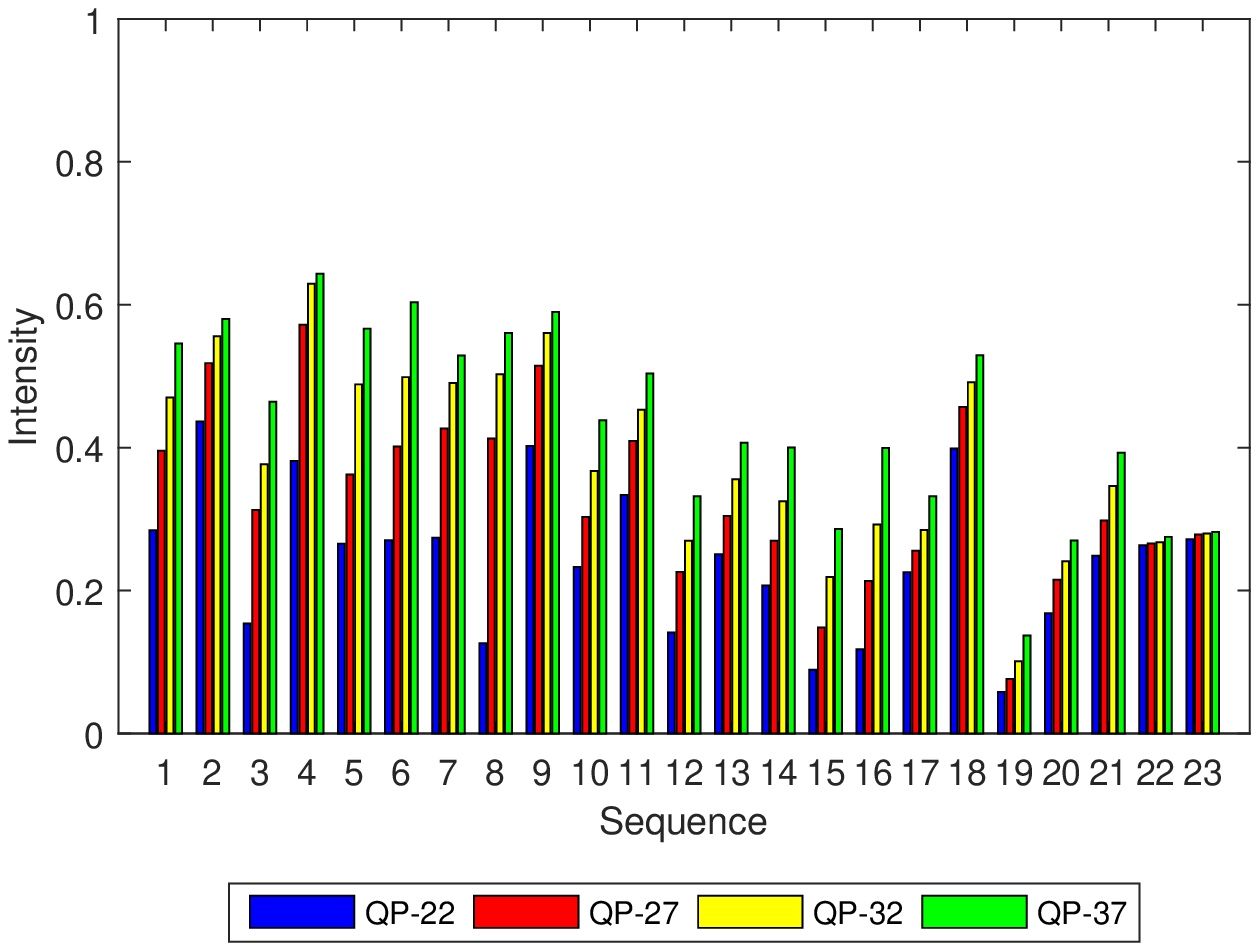}
  }
 \subfigure[Temporal PEA intensity]{
    \includegraphics[width=3.4in,height=6.5cm]{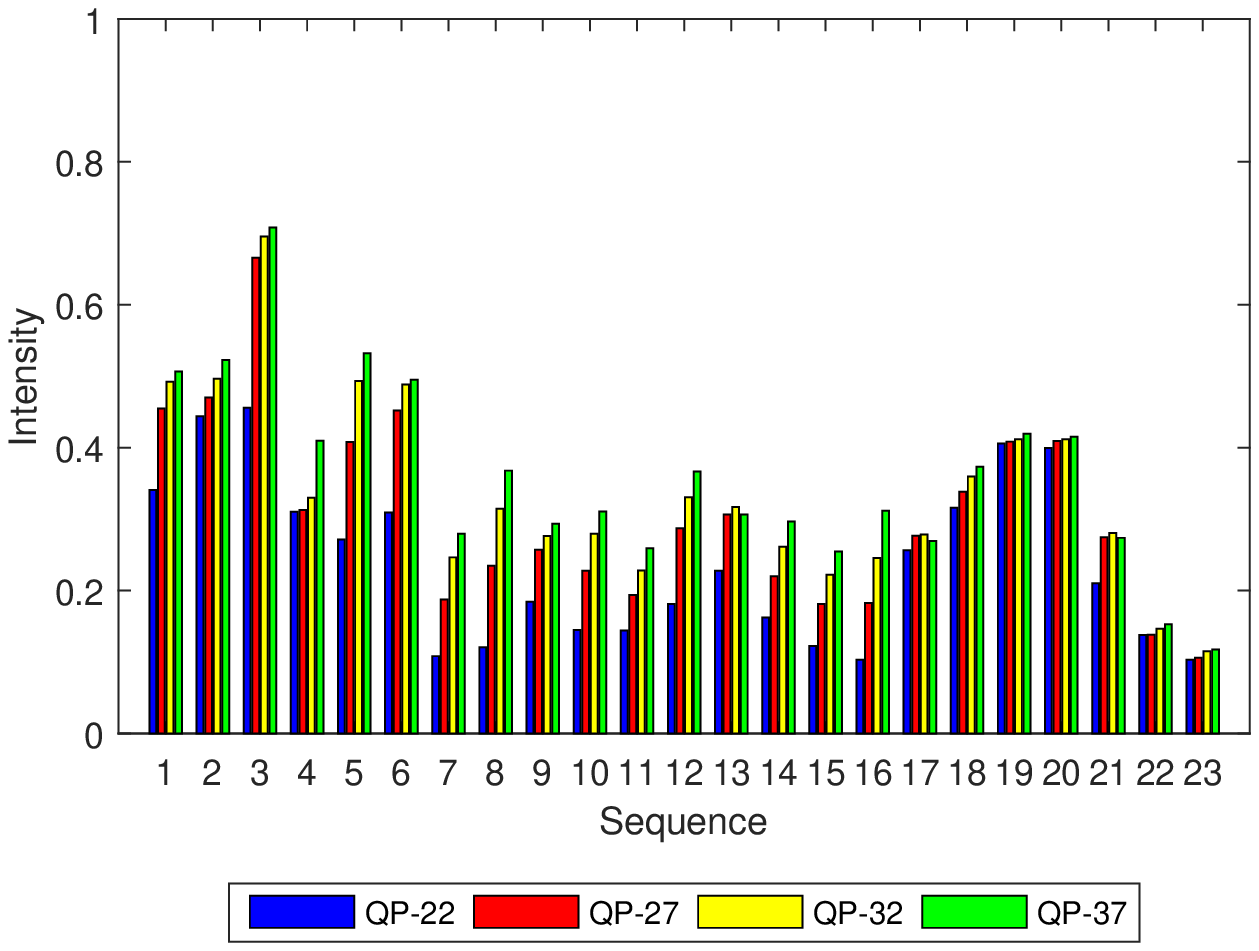}
  }
  \caption{The spatial and temporal PEA intensities on average.}
  \label{fig_PEA2_detection}
\end{figure*}

In order to further investigate the differences between spatial and temporal PEAs, we present the averaged PEA intensities for spatial and temporal artifacts in Fig. \ref{fig_PEA2_detection}. The aforementioned conclusions can also been verified in this figure.
\section{Conclusion}
We construct PEA265, a first-of-its-kind large-scale subject-labelled database of PEAs produced by H.265/HEVC video compression. The database contains 6 spatial and temporal PEA types, including blurring, blocking, ringing, color bleeding, flickering and floating, each with at least 60,000 samples with positive or negative labels. Using the database, we train CNNs to recognize PEAs, and the results show that state-of-the-art ResNext provides high accuracies in PEA detection. Moreover, we define a PEA intensity measure to assess the overall severeness of PEAs in compressed videos. This work will benefit the future development of video quality assessment algorithms. It can also be used to optimize hybrid video encoders for improved perceptual quality and perceptually-motivated video encoding schemes.

\bibliography{IEEE}
\flushend
\end{document}